\documentclass{article}

\PassOptionsToPackage{numbers, compress}{natbib}

\usepackage[preprint]{neurips_2026}
\usepackage[utf8]{inputenc} 
\usepackage[T1]{fontenc}    
\usepackage{hyperref}       
\usepackage{url}            
\usepackage{booktabs}       
\usepackage{amsfonts}       
\usepackage{amsmath}        
\usepackage{nicefrac}       
\usepackage{microtype}      
\usepackage{xcolor}         
\usepackage{graphicx}
\usepackage{float}
\usepackage{multirow}
\usepackage{enumitem}

\title{Unsupervised Adaptation of PDE Foundation Models}

\author{%
  Ziye Song \\
  Nanyang Technological University
  \And
  Zhao Wei \\
   CFAR,  A*STAR
  \AND
  Xin Yu \\
  Adelaide University
  \And
  Ivor Tsang \\
   CFAR,  A*STAR
  \And
  Yueming Lyu\thanks{\vspace{-20pt}Corresponding author: Yueming Lyu and Wei Zhao}~~ \\
  CFAR,  A*STAR
}

\begin{document}

\maketitle

\begin{abstract}
Pretrained partial differential equation (PDE) foundation models can generalize across different equations, but adapting them to unseen PDE systems typically requires dense solution data, which is often expensive or unavailable. To address this limitation, we propose an unsupervised PDE-based finetuning framework that eliminates the need for ground-truth solutions. We first pretrain a neighborhood attention Transformer on diverse time-dependent PDEs spanning varying spatial scales, yielding transferable representations across heterogeneous equations. In the adaptation stage, we construct a physics-based objective using the PDE residual and boundary conditions, and finetune the model on unseen equations via low-rank adaptation (LoRA). To address the uneven learning across physical quantities in standard LoRA, we introduce NSLoRA, a Newton-Schulz orthogonalized variant that rebalances adaptation. Our method achieves performance comparable to supervised LoRA finetuning without requiring any ground-truth solutions, while consistently outperforming competitive neural operator baselines and recent PDE foundation models across heterogeneous PDE benchmarks spanning multiple spatial dimensions. Code and models will be open-sourced.

\end{abstract}

\section{Introduction}
\label{sec:introduction}

Partial differential equations (PDEs) serve as the mathematical backbone of the physical sciences, with accurate and efficient PDE solvers being indispensable to a broad range of scientific and engineering applications. Traditional numerical methods, such as finite difference, finite element, and spectral methods~\cite{evans2010partial,leveque2007finite}, yield reliable accuracy at significant computational cost~\cite{karniadakis2021physics}, with high-quality scientific simulation often requiring specialized software and months of supercomputer time~\cite{ohana2024thewell}. To address these limitations, learning-based PDE solvers have emerged as a promising alternative. Neural operators~\cite{kovachki2023neural,li2021fourier,lu2021deeponet,li2023geo,wu2024transolver,wen2025gaot} learn mappings between function spaces from simulation data and achieve fast inference once trained. However, their performance critically depends on large-scale, high-fidelity datasets, where each training sample may itself require expensive numerical simulation~\cite{hao2024dpot,herde2024poseidon}. This data requirement makes them impractical in many real-world scenarios where simulations are costly or unavailable.

Physics-informed neural networks (PINNs)~\cite{raissi2019pinn} address this challenge by incorporating governing equations and initial/boundary conditions directly into the training objective, thereby reducing the need for labeled data. Despite this advantage, PINNs are typically optimized for individual PDE systems and can be difficult to train, often leading to suboptimal accuracy in practice~\cite{wang2020ntkpinn,wang2022causalpinn,wang2025gradientalignment,brandstetter2022lpsda}. To improve generalization, several works have explored meta-learning and multi-task extensions of PINNs, aiming to capture shared structures across related PDEs and enable faster adaptation to new tasks \cite{wei2026out}. While these approaches show improved efficiency within families of similar systems, their transferability remains limited when the underlying equations differ significantly. As a result, both neural operators and PINN-based methods still operate in a largely task-specific manner and struggle to generalize across heterogeneous PDEs.

Inspired by the success of foundation models in natural language processing~\cite{brown2020language,devlin2019bert} and computer vision~\cite{dosovitskiy2021vit}, recent works on PDE foundation models~\cite{mccabe2024mpp,herde2024poseidon,hao2024dpot,chen2025omniarch,holzschuh2025pdetransformer,nguyen2025physix} aim to learn representations shared across heterogeneous physical systems. By pretraining on diverse PDEs, these models can be adapted to new equations with improved accuracy. However, their adaptation stage remains largely data-driven, typically requiring dense ground-truth solution fields for fine-tuning. This limits their applicability in settings where such data is unavailable or expensive to obtain. In many practical scenarios, while labeled solution data is scarce, the governing equations and boundary conditions are readily available, yet existing approaches do not leverage this information during adaptation.


In this work, we propose an unsupervised PDE-based finetuning framework to address this downstream data dependency. To the best of our knowledge, our method is the first attempt to adapt a pretrained PDE foundation model to unseen equations strictly by enforcing the governing PDEs and boundary conditions. As a result, we can completely eliminate the requirement for interior ground truth data. 
To be specific, our framework operates in two stages: (i) a pretraining stage, which establishes transferable representation across heterogeneous equations, and (ii) an adaptation stage, which finetunes the model on a target equation under the unsupervised PDE-based adaptation objective.
Motivated by the demand to accommodate varying spatial resolutions without constructing resolution-specific positional encodings, we introduce neighborhood attention~\cite{hassani2023neighborhood} into the PDE foundation model setting to align with the localized nature of PDE dynamics. We initialize our framework by pretraining a backbone network on PDEBench~\cite{takamoto2022pdebench}, yielding highly transferable representation. This pretraining stage is essential, establishing the transferable representation that subsequent adaptation relies upon.

In the adaptation stage, we introduce the unsupervised PDE-based Adaptation Objective, which comprises a PDE residual term and a boundary condition term. The pretrained backbone remains frozen while low-rank adaptation (LoRA) modules~\cite{hu2022lora} are optimized under this objective. We observe that standard LoRA exhibits imbalanced learning across physical quantities, effectively adapting to some while leaving others under-optimized. To overcome this optimization bottleneck, we introduce NSLoRA, a variant that orthogonalizes each low-rank matrix via Newton-Schulz iteration~\cite{jordan2024muon}, $1.38\times$ faster end-to-end than SVD orthogonalization on these low-rank matrices. This formulation mitigates rank collapse and intrinsically improves the adaptation of the under-learned quantities. Collectively, these two stages enable our framework to generalize to heterogeneous unseen PDEs. Without using any interior labels, our method bounds the performance gap within a factor of 2.5 on seven of the eight 2D benchmarks compared to the supervised LoRA finetuning on the same backbone, and successfully outperforms at least one supervised neural operator baseline on 9 of 11 downstream datasets. Our contributions are summarized as follows:
\vspace{-0.5em}
\begin{itemize}[leftmargin=1.5em]
\item We propose the first unsupervised PDE-based finetuning paradigm that adapts a pretrained PDE foundation model to unseen equations relying exclusively on the governing PDEs and boundary conditions, thereby eliminating the requirement for any interior ground-truth solution fields.
\item We propose NSLoRA, a Newton-Schulz orthogonalized low-rank adapter that mitigates the rank collapse, deployed within a neighborhood attention backbone that natively accommodates inputs of varying spatial resolutions.
\item Under the unsupervised PDE-based Adaptation Objective, our framework achieves performance within a factor of $2.5$ relative to the supervised LoRA finetuning on seven of the eight heterogeneous 2D PDE benchmarks and outperforms at least one supervised neural operator baseline on $9$ of $11$ downstream datasets, despite no interior labels.
\end{itemize}

\begin{figure}[t]
\centering
\includegraphics[width=\textwidth]{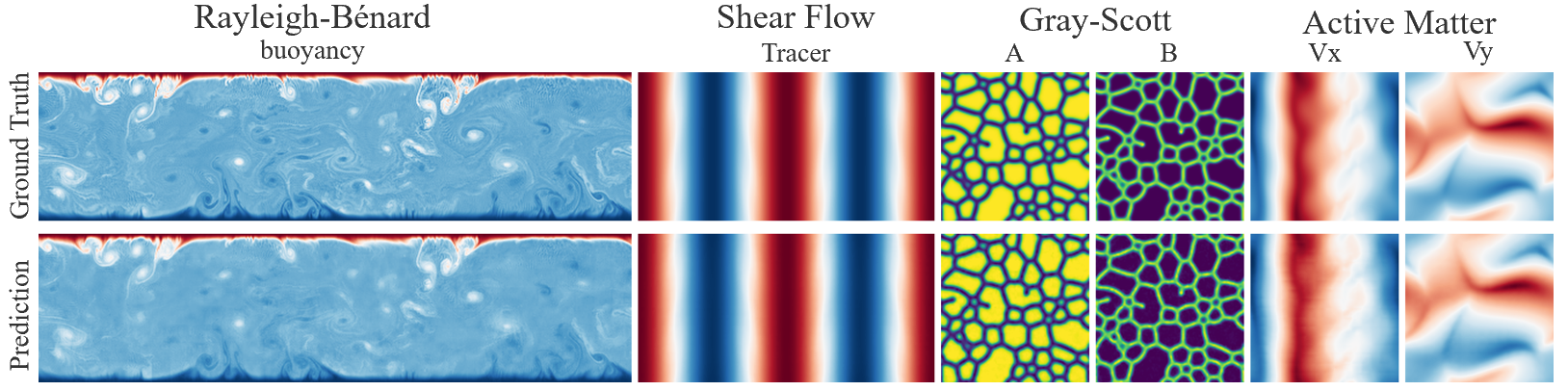}
\vspace{-1.5em}
\caption{Our unsupervised PDE-based finetuning framework predicts unseen PDEs strictly through the governing PDEs and boundary conditions, without any interior ground-truth fields. We visualize representative channels on the four datasets in The Well~\cite{ohana2024thewell}.}
\label{fig:present}
\vspace{-1.0em}
\end{figure}

\section{Related Work}
\label{sec:related}
\vspace{-0.5em}
\textbf{Neural operators.}
\citet{kovachki2023neural} laid the theoretical groundwork for approximating operators across infinite dimensional Banach spaces. FNO~\cite{li2021fourier} performs global convolution via truncated Fourier modes, while DeepONet~\cite{lu2021deeponet} factorizes the operator into input encoding and output reconstruction branches~\cite{chen1995universal}. Subsequent works extend these methodologies to irregular domains, factorized convolutions, and transformer-style attention over physical tokens~\cite{li2023geo,tran2023ffno,raonic2024cno,wu2024transolver,wen2025gaot}. Recent efforts address the task-specific constraint, with LeMON~\cite{sun2024lemon} training a single network across 19 distinct PDEs via meta-learning and \citet{serrano2026splitting} approximating unseen combined dynamics by operator splitting without retraining. All the preceding methods remain purely data-driven and rely on ground-truth solutions for training.

\textbf{Physics-informed methods.}
PINNs~\cite{raissi2019pinn,11454596} formulate the training loss as a combination of PDE residuals and initial/boundary conditions, reducing the reliance on dense solution data. Subsequent works address training pathologies from complementary angles such as training-dynamics analysis, temporal causality, gradient alignment, and symmetry-based data augmentation~\cite{wang2020ntkpinn,wang2022causalpinn,wang2025gradientalignment,brandstetter2022lpsda}. PINO~\cite{li2024pino} and PI-DeepONet~\cite{wang2021pideeponet} inject equation-level constraints into FNO and DeepONet respectively. Complementary to automatic differentiation, CAN-PINN~\cite{chiu2022can} replaces AD with grid-based stencils including second-order upwind schemes that our residual discretization in Section~\ref{sec:physlora} follows. Recent work begins to explore cross-PDE generalization within the physics-informed paradigm. HyPINO~\cite{bischof2025hypino_neurips} employs a hypernetwork conditioned on PDE parameters, and \citet{auroy2025perturbativepinn} pretrain a multi-head PINN on a linear operator and transfer to nonlinear PDEs in closed form without retraining. These works remain trained from random initialization and have not been coupled with pretrained foundation models.

\textbf{PDE foundation models and low-rank adaptation.}
Foundation models for PDEs are pretrained on heterogeneous PDE datasets~\cite{takamoto2022pdebench,ohana2024thewell,koehler2024apebench} and generalize to downstream tasks with significantly fewer labeled samples than training a specialized model from scratch. Poseidon~\cite{herde2024poseidon}, built upon a multiscale Swin Transformer, leverages the semi-group property of time-dependent PDEs. OmniArch~\cite{chen2025omniarch} accommodates variable physical quantities through a channel-wise state representation. PDE-Transformer~\cite{holzschuh2025pdetransformer} introduces a Diffusion Transformer backbone pretrained on APEBench~\cite{koehler2024apebench}. MORPH~\cite{rautela2025morph} supports heterogeneous spatial modalities and finetunes downstream tasks via LoRA~\cite{hu2022lora}, making it the closest in methodology to our work, yet it strictly relies on dense ground truth fields for adaptation. MPP~\cite{mccabe2024mpp} and PhysiX~\cite{nguyen2025physix} corroborate that joint pretraining over diverse physical systems facilitates transfer to unseen systems. Even where physics priors are incorporated, such as the PDE-Aligner module in OmniArch~\cite{chen2025omniarch} and the physics-informed temporal alignment in PITA~\cite{zhu2025pita}, these serve as auxiliary regularizers alongside the dominant supervised loss. PI-MFM~\cite{zhu2025pimfm} processes symbolic PDE representations and automatically assembles residual losses, enforcing physics constraints during both pretraining and adaptation across 1D time-dependent PDEs. Orthogonal in intent but complementary in level, recent LoRA variants~\cite{wang2023orthogonal,buyukakyuz2024olora,xiong2026oplora,wang2026orthogeolora} impose orthogonality at cross-task, initialization, projection, or Stiefel-manifold levels, and factor-wise Muon-style updates yield uniform spectral growth~\cite{jordan2024muon,kang2026uniform}. NSLoRA differs by applying Newton-Schulz in the forward pass with scalar magnitude parameters, and we detail this comparison in Section~\ref{sec:orthlora}.

\vspace{-0.5em}
\section{Approach}
\label{sec:approach}
\vspace{-0.5em}
\subsection{Problem Formulation}
\label{sec:formulation}
\vspace{-0.5em}
PDE problems are defined over a domain $\Omega$ with the corresponding initial conditions and boundary conditions. Our goal is to adapt a pretrained PDE foundation model to unseen PDEs relying exclusively on these physical constraints, thereby eliminating the need for interior ground-truth solution fields required by conventional supervised finetuning.

We consider a generalized spatiotemporal physical system, where a $c$-channel multivariate physical field $\mathbf{u}: \Omega \times [0, T] \to \mathbb{R}^c$ on a $d$-dimensional domain $\Omega \subset \mathbb{R}^d$ ($d \in \{1, 2, 3\}$) satisfies:
\begin{equation}
\begin{aligned}
\partial_t \mathbf{u}(\mathbf{x}, t) + \mathcal{N}[\mathbf{u}(\mathbf{x}, t)] &= 0, &\quad \mathbf{x} &\in \Omega,\ t \in (0, T], \\
\mathbf{u}(\mathbf{x}, 0) &= \mathbf{u}_0(\mathbf{x}), &\quad \mathbf{x} &\in \Omega, \\
\mathcal{B}[\mathbf{u}(\mathbf{x}, t)] &= g(\mathbf{x}, t), &\quad \mathbf{x} &\in \partial\Omega,\ t \in (0, T],
\end{aligned}
\label{eq:pde_system}
\end{equation}
where  $\mathcal{N}$ denotes a general (spatial) differential operator encapsulating potentially nonlinear spatial dynamics, $\mathbf{u}_0$ specifies the initial condition at $t = 0$, and $\mathcal{B}$ is the prescribed boundary operator that enforces the boundary value $g$ on $\partial\Omega$. We further denote the residual operator as $R(\mathbf{u}) := \partial_t \mathbf{u} + \mathcal{N}[\mathbf{u}]$, so that any solution of the governing equation satisfies $R(\mathbf{u}) \equiv 0$ on $\Omega$ and $R(\hat{\mathbf{u}})$ quantifies the pointwise violation of the equation by a candidate prediction $\hat{\mathbf{u}}$.

We formulate PDE solving as next-step operator learning. Given $T_{\mathrm{in}}$ temporally contiguous frames $\mathbf{U}_t = [\mathbf{u}_{t - T_{\mathrm{in}} + 1}, \ldots, \mathbf{u}_t]$, the model learns a surrogate mapping $\mathcal{G}_\theta$ that predicts the immediate subsequent state $\hat{\mathbf{u}}_{t+1} = \mathcal{G}_\theta(\mathbf{U}_t)$. $\mathcal{G}_\theta$ underlies both pretraining and downstream adaptation. This formulation addresses deployment settings in which a short history of the target system is already available, such as extending or accelerating an existing coarse simulation, or continuing a trajectory from recorded observations. It does not address the cold-start setting in which only an initial condition and a set of equation parameters are given, which would require the model to produce the first $T_{\mathrm{in}}$ frames itself.

\noindent\textit{Unsupervised PDE-Based Adaptation Objective (UPAO).} Given a pretrained operator $\mathcal{G}_{\theta_0}$, an unseen PDE specified by $\mathcal{N}$, an initial history window $\mathbf{U}_t$, and a boundary reference field $\mathbf{u}^*$ defined on $\partial\Omega \times [0, T]$, we seek to optimize parameters $\theta^*$ such that $\mathcal{G}_{\theta^*}$ approximates the solution operator on $\Omega$ for $t > 0$, without access to the ground truth solution at any interior point. Throughout the paper, $\mathbf{u}^*$ exclusively denotes this boundary reference field, which is instantiated as the peripheral band of width $1$ grid points along the outermost spatial boundary, remaining intrinsically accessible across all temporal steps. We detail the corresponding training objective in Section~\ref{sec:physlora}.

\vspace{-0.75em}
\subsection{Architecture}
\label{sec:architecture}
\vspace{-0.5em}
\begin{figure}[t]
\centering
\includegraphics[width=\textwidth]{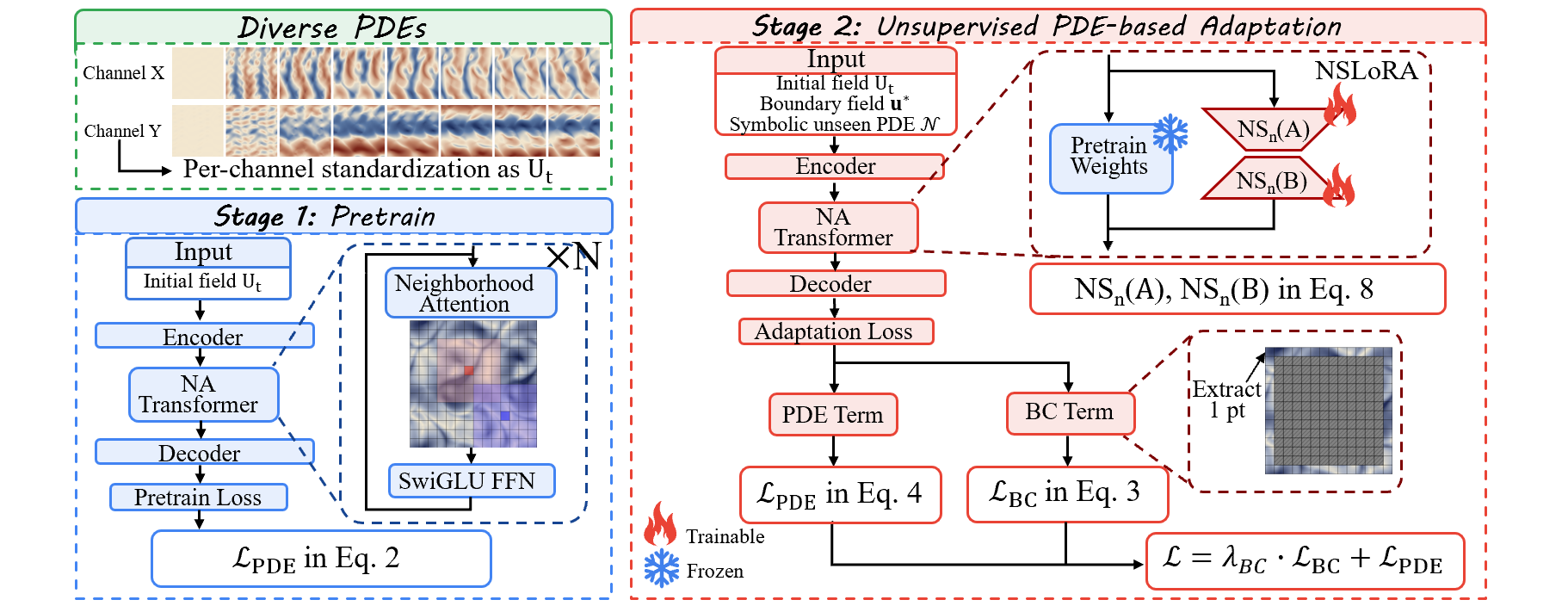}
\vspace{-1.5em}
\caption{Overview of our framework. The shared backbone $\mathcal{G}_\theta$ inherently accommodates inputs of varying spatial resolutions. The encoder partitions each input frame into patches and projects them into a token sequence, while the decoder inverts the patchification. A stack of neighborhood attention layers processes the tokens. Stage 1 pretrains the backbone via supervised next-step prediction. Stage 2 strictly freezes the pretrained weights and inserts NSLoRA adapters into every attention and feed-forward module, optimized under our unsupervised PDE-based adaptation objective.}
\label{fig:overview}
\vspace{-0.5em}
\end{figure}

Figure~\ref{fig:overview} demonstrates the overall architecture of our proposed model. We employ a neighborhood attention (NA) Transformer~\cite{hassani2023neighborhood} as the backbone for $\mathcal{G}_\theta$, whose window-restricted attention inherently aligns with the localized nature of PDE dynamics and scales linearly with the sequence length given a fixed window size. The NA Transformer leverages a patch-based encoder-decoder paradigm with a Transformer core applied uniformly across spatial domains, strictly preserving spatiotemporal dimensionality between input and output. Heterogeneous physical fields are unified into a shared $C$-channel representation. Given the input $\mathbf{U}_t$, the model generates a temporally shifted sequence $\hat{\mathbf{U}}_{t+1} = [\hat{\mathbf{u}}_{t-T_{\mathrm{in}}+2}, \ldots, \hat{\mathbf{u}}_{t+1}]$ of equivalent length, from which only the last frame $\hat{\mathbf{u}}_{t+1}$ is retained as the next-step prediction. This shifted predictive design imposes dense supervision across every temporal subsequence from $1$ to $T_{\mathrm{in}}$ within a single forward, enabling inference with  arbitrary temporal length up to $T_{\mathrm{in}}$ frames without structural alterations.

\textbf{Unified channel representation.} Motivated by the necessity to process diverse physical fields from heterogeneous datasets, the unified architecture is engineered to accommodate flexible input channel compositions. Each physical field is projected into a unified $C$-dimensional state space, with specific indices deterministically assigned to velocity channels and scalar channels according to the distinct configuration of each dataset (detailed in Appendix~\ref{app:slot_mapping}). Specifically, a binary mask selectively activates the required channels while strictly zero-padding the inactive channels.

\textbf{Encoder-decoder.} The encoder $\mathcal{E}$ and decoder $\mathcal{D}$ constitute a symmetric patch-based architecture. To accommodate the varying magnitude scales across distinct physical quantities, the input $\mathbf{U}_t$ is strictly subjected to a preliminary standardization $\tilde{\mathbf{U}}_t = (\mathbf{U}_t - \mu)/(\sigma + \epsilon)$, where $\mu$ and $\sigma$ represent the mean and standard deviation computed independently for each sample and physical quantity of $\mathbf{U}_t$, ensuring that all subsequent internal representations operate in normalized space. Given $\tilde{\mathbf{U}}_t \in \mathbb{R}^{T_{\mathrm{in}} \times N_1 \times \cdots \times N_d \times C}$ on a spatial grid of resolution $N_1 \times \cdots \times N_d$, the encoder partitions each frame into non-overlapping $d$-dimensional patches of side length $p$ and yields a token sequence $\mathbf{Z} \in \mathbb{R}^{T_{\mathrm{in}} \times M \times H}$ with embedding dimension $H$ and $M = \prod_{i=1}^d (N_i/p)$ patches per frame. Each token $\mathbf{z}_{t,m} \in \mathbb{R}^H$, representing the corresponding spatiotemporal volume of $\mathbf{U}_t$, is computed through a stack of strided convolutions with residual blocks that downsamples the patch to an $H$-dimensional vector. The decoder $\mathcal{D}$ executes the inverse mapping of this encoding process. Transposed convolutions upsample each token to its $d$-dimensional patch, and the patches are rearranged into a full-resolution intermediate field. A convolution formulated as $\mathbf{I} + \mathrm{Conv}_0$ refines $\tilde{\mathbf{u}}_{t+1}$ to mitigate the boundary reconstruction artifacts. Finally, the inverse transform $\hat{\mathbf{u}}_{t+1} = \sigma \cdot \tilde{\mathbf{u}}_{t+1} + \mu$ restores the prediction to its original physical scale.

\textbf{Shared NA Transformer.} The Transformer core $\mathcal{T}$ processes the token sequence $\mathbf{Z} \in \mathbb{R}^{T_{\mathrm{in}} \times M \times H}$ through $L$ pre-norm residual layers. Each layer contains a dimension-specific neighborhood attention $\mathrm{NA}^{(d)}$ and a shared SwiGLU~\cite{shazeer2020glu} feed-forward block. The feed-forward block $\mathrm{SwiGLU}(\mathbf{z}) \in \mathbb{R}^H$ and the preceding RMSNorm~\cite{zhang2019rmsnorm} are applied pointwise to each token. Therefore their learnable weights are universally shared across any dimension $d$. $\mathrm{NA}^{(d)}$ constrains each token at position $(t, x_1, \ldots, x_d)$ to a local window of width $k$ along each axis, totalling $k^{d+1}$ tokens that span one temporal and $d$ spatial axes. Consequently, the attention kernel strictly imposes distinct instantiations and unshared weights across different spatial dimensionalities. $\mathcal{T}$ evaluates to $O(L \cdot M \cdot T_{\mathrm{in}} \cdot k^{d+1})$, exhibiting linear complexity scaling with the token count under a fixed $k$.

Each NA variant enforces causality along the temporal axis while remaining unconstrained along the spatial axes, an asymmetry strictly mandated by the shifted predictive design introduced above. Specifically, since the input $\mathbf{u}_{\tau+1}$ inherently encapsulates the exact supervision target (for $\tau < T_{\mathrm{in}} - 1$), the predicted output $\hat{\mathbf{u}}_{\tau+1}$ would directly exploit the value of $\mathbf{u}_{\tau+1}$ under an unmasked temporal attention, degenerating the learning objective into a trivial identity mapping $\mathbf{u}_{\tau+1} \mapsto \mathbf{u}_{\tau+1}$. Temporal causality strictly precludes such information leakage. Spatial attention remains bidirectional, since spatial dimensions do not necessitate such causality. 

\textbf{Supervised pretraining.} We pretrain the operator $\mathcal{G}_\theta$ on heterogeneous time-dependent PDEs through supervised next-step prediction. Given a history window $\mathbf{U}_t$ paired with its ground-truth shifted counterpart $\mathbf{U}_{t+1} = [\mathbf{u}_{t - T_{\mathrm{in}} + 2}, \ldots, \mathbf{u}_{t+1}]$, we minimise the mean-squared discrepancy across every output frame. The pretraining loss is formulated as:
\begin{equation}
\mathcal{L}_{\mathrm{pre}} = \frac{1}{T_{\mathrm{in}}\,|\Omega|} \big\| \hat{\mathbf{U}}_{t+1} - \mathbf{U}_{t+1} \big\|^2,
\label{eq:pretrain_loss}
\end{equation}
where the computation is restricted to the active channels. Through pretraining, the resulting backbone $\mathcal{G}_{\theta_0}$ encodes the transferable representation that the adaptation stage subsequently leverages.

\vspace{-0.75em}
\subsection{Unsupervised PDE-Based Adaptation}
\label{sec:physlora}
\vspace{-0.5em}

The UPAO, or Unsupervised PDE-Based Adaptation Objective, is formulated through a dual-component training objective. Given an initial history window $\mathbf{U}_t$, we partition the spatial domain into a boundary band $\mathcal{S}_b$ and an interior region $\mathcal{S}_p$, with $\mathcal{S}_p \cup \mathcal{S}_b = \Omega$ and $\mathcal{S}_p \cap \mathcal{S}_b = \emptyset$. The boundary condition term $\mathcal{L}_{\mathrm{BC}}$ supervises $\mathcal{S}_b$ by aligning the prediction with the boundary reference field $\mathbf{u}^*$, while the PDE residual term $\mathcal{L}_{\mathrm{PDE}}$ penalises the equation residual $R(\hat{\mathbf{u}})$ on $\mathcal{S}_p$ through finite-difference discretization.

\textbf{Boundary condition term.}
The reference field $\mathbf{u}^*$ in $\mathcal{L}_{\mathrm{BC}}$ constitutes the boundary observation specified by the UPAO, evaluated on a peripheral band of width $1$ grid points along the spatial boundary. We supervise $\hat{\mathbf{u}}$ directly against $\mathbf{u}^*|_{\partial\Omega}$ within this band. For each predicted frame, with $\mathcal{S}_b$ denoted as the set of boundary grid points within the band, we compute the MSE between the prediction and the reference within this region. This boundary loss is formulated as:
\begin{equation}
\mathcal{L}_{\mathrm{BC}} = \sqrt{\frac{1}{|\mathcal{S}_b|} \sum_{\mathbf{x} \in \mathcal{S}_b} \|\hat{\mathbf{u}}_{t+1}(\mathbf{x}) - \mathbf{u}^*_{t+1}(\mathbf{x})\|^2} \,.
\label{eq:bc_loss}
\end{equation} 
\textbf{PDE residual term.}
We evaluate the discrete residual $R(\hat{\mathbf{u}})$ on every grid point of $\mathcal{S}_p$ through finite difference schemes~\cite{leveque2007finite}. Diffusive and gradient operators use central differences, while convective nonlinearities such as those in Navier--Stokes or Burgers employ a conservative upwind discretization following~\cite{chiu2022can}. Implementation specifics, including ghost-cell extrapolation and the upwind face-value formulas, are deferred to Appendix~\ref{app:PDEresidual}.

Let $R_k$ denote the residual of the $k$-th sub-equation of the system. As the system decomposes into $K$ sub-equations, the corresponding residual scales can span multiple orders of magnitude. To mitigate this disparity, we apply an equation-wise normalization for each predicted frame. The normalized PDE loss is formulated as:
\begin{equation}
\mathcal{L}_{\mathrm{PDE}} = \sqrt{\frac{1}{K}\sum_{k = 1}^{K} \frac{1}{s_k^2 \, |\mathcal{S}_p|} \sum_{\mathbf{x} \in \mathcal{S}_p} \|R_k(\hat{\mathbf{u}})(\mathbf{x})\|^2} \,,
\label{eq:pde_loss}
\end{equation}
where $s_k^2 = \frac{1}{T_{\mathrm{in}}\,|\mathcal{S}_p|} \|R_k(\mathbf{U}_t)\|^2$ denotes the magnitude evaluated on $\mathbf{U}_t$ for the $k$-th sub-equation. This per-equation rescaling prevents any single sub-equation from dominating the gradient signal.

\textbf{Total loss.}
The overall optimization objective is defined as:
\begin{equation}
\mathcal{L} = \lambda_{\mathrm{BC}} \cdot \mathcal{L}_{\mathrm{BC}} + \mathcal{L}_{\mathrm{PDE}} \,.
\label{eq:total_loss}
\end{equation}
The scale of $\mathcal{L}_{\mathrm{PDE}}$ exhibits significant variance across heterogeneous PDE benchmarks. To address this instability, we designate $\lambda_{\mathrm{BC}}$ as the sole hyperparameter regulating the relative weight between $\mathcal{L}_{\mathrm{BC}}$ and $\mathcal{L}_{\mathrm{PDE}}$ during adaptation.

\vspace{-0.75em}
\subsection{Adapting the Pretrained Backbone via NSLoRA}
\label{sec:orthlora}
\vspace{-0.5em}
Under UPAO, we finetune the pretrained backbone $\mathcal{G}_{\theta_0}$ on each unseen target PDE by injecting low-rank adaptation (LoRA)~\cite{hu2022lora} into every NA layer and SwiGLU block, with the NA Transformer core strictly frozen at its pretrained values while the encoder and decoder remain jointly trainable. Standard LoRA freezes each pretrained weight matrix $W_0 \in \mathbb{R}^{d_{\mathrm{out}} \times d_{\mathrm{in}}}$ inherited from $\theta_0$ and incorporates a trainable rank-$r$ increment modulated by the fixed scalar $\alpha / r$, yielding the effective weight $W = W_0 + \Delta W$ with $r \ll \min(d_{\mathrm{in}}, d_{\mathrm{out}})$. This increment is defined as:
\begin{equation}
\Delta W = \frac{\alpha}{r}\, B A, \quad A \in \mathbb{R}^{r \times d_{\mathrm{in}}}, \; B \in \mathbb{R}^{d_{\mathrm{out}} \times r} \,.
\label{eq:lora}
\end{equation}
The bottleneck dimension $r$ of standard LoRA strictly upper-bounds the effective rank of $\Delta W$, defined as the number of nontrivial singular values. During adaptation, this effective rank frequently collapses because no constraint prevents redundancy among the columns of $B$ and rows of $A$. The resulting rank deficiency restricts the adaptation of less dominant physical quantities. Enforcing orthogonality on each low-rank matrix ensures that every rank component is a unit-norm direction, increasing the effective rank and balancing the adaptation across the physical quantities.

We propose NSLoRA, which applies Newton-Schulz (NS) iteration~\cite{jordan2024muon} to each low-rank matrix independently and rescale them by the original Frobenius norms. This increment is formulated as:
\begin{equation}
\Delta W = \|A\|_F \cdot \|B\|_F \cdot \mathrm{NS}(B)\, \mathrm{NS}(A) \,,
\label{eq:orthlora}
\end{equation}
where $\mathrm{NS}(B)$ returns an approximately column-orthogonal matrix and $\mathrm{NS}(A)$ returns an approximately row-orthogonal matrix. Both matrices are approximated via $n$ NS steps that iteratively apply
\begin{equation}
X \leftarrow a\, X + b\, (X X^{\top}) X + c\, (X X^{\top})^2 X \,,
\label{eq:ns_iter}
\end{equation}
with predefined coefficients $(a, b, c)$ inherited from~\cite{jordan2024muon}. Since $\|A\|_F$ and $\|B\|_F$ are computed from the learnable matrices at every forward pass, the rescaling $\|A\|_F \cdot \|B\|_F$ replaces the fixed $\alpha / r$ in Eq.~\ref{eq:lora} with an adaptive per-module amplitude. 

\section{Experiments}
\label{sec:experiments}
\vspace{-0.5em}
\subsection{Setup}
\label{sec:setup}
\vspace{-0.5em}
\textbf{Datasets.}
We pretrain the backbone on six PDEBench subsets~\cite{takamoto2022pdebench} covering compressible Navier-Stokes in 1D, 2D, and 3D, shallow water, diffusion-reaction, and incompressible Navier-Stokes. We evaluate UPAO on eleven downstream datasets across two suites. The first suite contains four high-fidelity datasets from The Well~\cite{ohana2024thewell}, including Rayleigh--B\'enard, Shear Flow, Gray--Scott reaction-diffusion, and Active Matter. The second suite contains seven datasets analytically derived from exact solutions, covering Burgers, Advection (Adv), Taylor--Green (TG), Wave, and Advection--Diffusion (AdvDiff) equations across 1D, 2D, and 3D. Further details on these learning tasks are deferred to Appendix~\ref{app:datasets}. The selection of datasets is strictly motivated by the formulation of our UPAO objective. Since UPAO relies on finite difference approximations to compute the PDE residual term, it is highly sensitive to the numerical precision of the data. Certain simulation benchmarks, such as PDEArena~\cite{gupta2022towards} and PDEGym~\cite{herde2024poseidon}, often introduce discretization errors that severely compromise the accuracy of computed PDE residuals. Therefore, we utilize datasets synthesized from exact analytical solutions, which guarantee absolute precision at discrete coordinates. Detailed visualizations of the computed PDE residuals are provided in Figure~\ref{fig:pde_residual}. Crucially, to assess cross-equation transferability, seven of these eleven tasks involve unseen PDEs during pretraining. 

\textbf{Models and baselines.}
We compare against four neural-operator baselines: FNO~\cite{li2021fourier}, TFNO~\cite{kossaifi2023multigrid}, U-Net~\cite{ronneberger2015unet}, and CNextU-Net~\cite{liu2022convnext}, all trained from random initialization. Two PDE foundation models, PDE-Transformer~\cite{holzschuh2025pdetransformer} and Poseidon~\cite{herde2024poseidon}, are finetuned from publicly released checkpoints of a comparable scale to our backbone, ensuring a fair comparison. We additionally finetune PDE-Transformer under UPAO. Our backbone is evaluated under two finetuning objectives: a supervised LoRA which serves as an empirical upper bound, and UPAO with NSLoRA. Our backbone consumes a history window of $T_{\mathrm{in}} = 8$ frames and predicts the next frame, with the same window length used in pretraining, adaptation, and evaluation. The two foundation-model baselines ingest temporal context differently, as detailed in Appendix~\ref{app:fm_baselines}. Complete implementation details for all baselines can be found in Appendix~\ref{app:baselines}.

\textbf{Evaluation metric.}
Following~\cite{ohana2024thewell}, we report the variance-normalized root mean squared error (VRMSE) averaged across the $c$ active channels of each dataset:
\begin{equation}
\mathrm{VRMSE}(\hat{\mathbf{u}}, \mathbf{u}) = \frac{1}{c} \sum_{i=1}^{c} \sqrt{\frac{\sum_{\mathbf{x}, t} (\hat{u}_i - u_i)^2}{\sum_{\mathbf{x}, t} (u_i - \bar{u}_i)^2}},
\label{eq:vrmse_main}
\end{equation}
where $\hat{u}_i$ and $u_i$ denote the $i$-th channels of $\hat{\mathbf{u}}$ and $\mathbf{u}$, respectively, and $\bar{u}_i$ is the temporal mean of $u_i$ at each spatial location. A VRMSE above $1$ indicates worse results than an accurate estimation of $\bar{u}_i$.

\subsection{Main Results}
\label{sec:main_results}
\vspace{-0.5em}
\begin{table}[t]
\centering
\caption{Evaluation on eleven downstream datasets, measured by VRMSE ($\downarrow$). \textbf{sup}: supervised training with dense ground-truth labels. \textbf{unsup}: training with only boundary observations and PDE residuals (UPAO). \textbf{NO}: Neural Operator. \textbf{FM}: Foundation Model. PDE-Trans. abbreviates PDE-Transformer.}
\label{tab:main}
\resizebox{\textwidth}{!}{%
\begin{tabular}{c c l cccc ccccccc}
\toprule
& & & \multicolumn{4}{c}{\textbf{The Well~\cite{ohana2024thewell}}} & \multicolumn{7}{c}{\textbf{Exact-Solution Datasets}} \\
\cmidrule(lr){4-7}\cmidrule(lr){8-14}
& & Method
  & \shortstack{Rayleigh\\B\'{e}nard}
  & \shortstack{Shear\\Flow}
  & \shortstack{Gray\\Scott}
  & \shortstack{Active\\Matter}
  & \shortstack{Burgers\\1D}
  & \shortstack{Adv\\1D}
  & \shortstack{TG\\ 2D}
  & \shortstack{Wave\\2D}
  & \shortstack{AdvDiff\\2D}
  & \shortstack{Burgers\\2D}
  & \shortstack{Adv\\3D} \\
\midrule
\multirow{8}{*}{sup} 
& \multirow{4}{*}{NO} & FNO~\cite{li2021fourier}              & 0.2139          & 0.3782          & 0.0073          & 0.1584          & 0.0063          & 0.0062          & 0.0029          & 0.0275          & 0.0296          & 0.1875          & 0.0645 \\
&                     & TFNO~\cite{kossaifi2023multigrid}     & 0.1762          & 0.2234          & \textbf{0.0033} & 0.1390          & \textbf{0.0009} & \textbf{0.0017} & 0.0033          & \textbf{0.0074} & 0.0040          & 0.1615          & 0.0197 \\
&                     & U-Net~\cite{ronneberger2015unet}      & 0.1423          & 0.2439          & 0.0182          & 0.2133          & 0.0455          & 0.0662          & 0.0197          & 0.0434          & 0.0357          & 0.2066          & 0.0570 \\
&                     & CNextU-Net~\cite{liu2022convnext}     & \textbf{0.1022} & 0.0455          & 0.0054          & 0.1248          & 0.0065          & 0.0088          & 0.0178          & 0.0082          & 0.0062          & 0.0165          & \textbf{0.0166} \\
\cmidrule(lr){2-14} 
& \multirow{3}{*}{FM} & PDE-Trans.~\cite{holzschuh2025pdetransformer} & 0.5320  & 0.0922          & 0.0876          & 0.0928          & ---             & ---             & 0.0159          & 0.0333          & 0.0149          & 0.0224          & --- \\
&                     & Poseidon~\cite{herde2024poseidon}   & 0.4946          & \textbf{0.0308} & 0.1845          & 0.4540          & ---             & ---             & 0.0473          & 0.1935          & 0.0146          & 0.0504          & --- \\
&                     & Ours                                  & 0.1660          & 0.0446          & 0.0754          & \textbf{0.0639} & 0.0112          & 0.0126          & \textbf{0.0014} & 0.0097          & \textbf{0.0023} & \textbf{0.0022} & 0.0979 \\
\midrule
\multirow{2}{*}{unsup}
& \multirow{2}{*}{FM} & PDE-Trans.~\cite{holzschuh2025pdetransformer} & 1.6302  & 1.4729          & 0.2009          & 1.3398          & ---             & ---             & 0.0137          & 0.1806          & 0.0418          & 0.2524          & --- \\
&                     & Ours                                  & \textbf{0.1812} & \textbf{0.1112} & \textbf{0.1099} & \textbf{0.1215} & \textbf{0.0142} & \textbf{0.0191} & \textbf{0.0060} & \textbf{0.0218} & \textbf{0.0038} & \textbf{0.0021} & \textbf{0.0985} \\
\bottomrule
\end{tabular}%
}
\end{table}

\textbf{Per-dataset comparison.}
In Table~\ref{tab:main}, our framework reduces VRMSE by a geometric-mean factor of $9.9$ compared to PDE-Transformer with standard LoRA under identical UPAO. Compared to the supervised LoRA upper bound on the same backbone, our unsupervised framework remains within $2.5\times$ on seven of the eight 2D benchmarks. Against the four competitive neural operator baselines, our framework outperforms at least one on nine of the eleven datasets. These results validate that UPAO effectively bridges the performance gap between unsupervised adaptation and fully supervised paradigms.

\textbf{Cross-target transfer and backbone dependence.}
\label{sec:cross_domain}
Table~\ref{tab:cross_domain} evaluates cross-target transferability.
While the main results establish the broad adaptation capabilities of foundation models, Table~\ref{tab:cross_domain} complements this evaluation by investigating the cross-dataset transferability of  neural operators. Each row corresponds to a source dataset on which the model is trained, and each column corresponds to a target dataset on which the model is evaluated. The diagonal entries represent the fully supervised in-domain references trained and evaluated on identical datasets. Off-diagonal cells assess transferability by finetuning a trained source model on a distinct target domain via UPAO. Our zero-shot transfer from the PDEBench-pretrained backbone is stronger than every CNextU-Net source checkpoint on Active Matter and Gray--Scott and weaker on the remaining five targets, and the subsequent UPAO finetuning reduces the target VRMSE by 1.4x to 7.0x on the four benchmarks from The Well. Conversely, applying UPAO to CNextU-Net~\cite{liu2022convnext} yields marginal improvements or noticeable performance degradation. These finetuning errors consistently remain at least an order of magnitude worse than their corresponding in-domain references, indicating that CNextU-Net achieves noticeably smaller transfer gains under the same UPAO objective than our framework. The magnitude of the UPAO gain is nevertheless backbone-dependent rather than exclusive to our architecture. On CNextU-Net, UPAO improves over zero-shot transfer on $14$ of the $26$ comparable off-diagonal cells and degrades it on the remaining $12$. On TFNO, it improves all $26$ cells, as reported in Appendix~\ref{app:cross_domain_full}. Taking the best UPAO result per target across source checkpoints, our pretrained backbone attains the lowest VRMSE on six of the seven targets, with Gray--Scott the exception, where the global Fourier modes of TFNO match the low-wavenumber pattern structure more efficiently than a $5 \times 5$ neighborhood window.

\begin{table}[t]
\centering
\caption{Evaluation of cross-dataset transfer under UPAO finetuning and zero-shot inference, measured by target VRMSE ($\downarrow$). Within each $7\times7$ panel, bold marks the lowest off-diagonal value per \emph{column}, i.e.\ the source dataset that transfers best to that target. Gray diagonal entries denote the in-domain supervised performance and are excluded from the comparison. In the two ``Ours'' columns the row index denotes the \emph{target} dataset and the source is always the PDEBench-pretrained backbone, so these columns have no diagonal. Bold there marks the better of zero-shot and UPAO for that target. ``---'' marks incompatible channel counts in the zero-shot setting. Dataset abbreviations: BG (Burgers 2D), WV (Wave 2D), TG (Taylor--Green 2D), AM (Active Matter), GS (Gray--Scott), RB (Rayleigh--B\'enard), SF (Shear Flow).}
\label{tab:cross_domain}
\resizebox{\textwidth}{!}{
\setlength{\tabcolsep}{3pt}
\begin{tabular}{l | ccccccc | ccccccc | cc}
\toprule

\multicolumn{1}{l}{} & \multicolumn{14}{c}{\textbf{CNextU-Net~\cite{liu2022convnext}}} & \multicolumn{2}{c}{\textbf{(c) Ours}} \\

\multicolumn{1}{l}{} & \multicolumn{7}{c}{\textbf{(a) Zero-shot}} & \multicolumn{7}{c}{\textbf{(b) UPAO}} & \multirow{2}{*}{\textbf{Zero-shot}} & \multirow{2}{*}{\textbf{UPAO}} \\
\cmidrule(lr){2-8} \cmidrule(lr){9-15}

\multicolumn{1}{l}{} & BG & WV & TG & AM & GS & RB & \multicolumn{1}{c}{SF} & BG & WV & TG & AM & GS & RB & \multicolumn{1}{c}{SF} & & \\
\midrule
BG & \textcolor{gray}{0.0165} & 0.2440 & --- & --- & 0.2576 & --- & --- & \textcolor{gray}{0.0165} & 0.3200 & 0.4098 & 1.5086 & 0.2424 & 0.5900 & 0.5383 & 0.0876 & \textbf{0.0021} \\
WV & 0.0845 & \textcolor{gray}{0.0082} & --- & --- & 0.2900 & --- & --- & \textbf{0.0200} & \textcolor{gray}{0.0082} & 0.3614 & 0.9588 & 0.3419 & 0.5854 & 0.6025 & 1.0176 & \textbf{0.0218} \\
TG & 0.3129 & 0.3522 & \textcolor{gray}{0.0178} & 0.8263 & 0.2918 & --- & --- & 0.1960 & 0.3582 & \textcolor{gray}{0.0178} & 0.6888 & \textbf{0.1456} & 0.5779 & 0.4282 & 0.1274 & \textbf{0.0060} \\
AM & \textbf{0.0555} & \textbf{0.0676} & \textbf{0.0485} & \textcolor{gray}{0.1248} & 0.2894 & --- & --- & 0.0367 & 0.1769 & \textbf{0.0474} & \textcolor{gray}{0.1248} & 0.2188 & \textbf{0.4404} & 0.4802 & 0.1935 & \textbf{0.1215} \\
GS & 0.4596 & 0.3965 & --- & --- & \textcolor{gray}{0.0054} & --- & --- & 0.2828 & 0.1657 & 0.3729 & 0.9479 & \textcolor{gray}{0.0054} & 0.6455 & 0.5094 & 0.1567 & \textbf{0.1099} \\
RB & 0.1197 & 0.1550 & 0.1561 & \textbf{0.3185} & 0.4879 & \textcolor{gray}{0.1022} & \textbf{0.2017} & 0.0604 & \textbf{0.1314} & 0.0624 & \textbf{0.3673} & 0.2701 & \textcolor{gray}{0.1022} & \textbf{0.2318} & 0.7068 & \textbf{0.1812} \\
SF & 0.0629 & 0.2403 & 0.0635 & 0.7857 & \textbf{0.2486} & \textbf{0.4492} & \textcolor{gray}{0.0455} & 0.2094 & 0.3964 & 0.1703 & 0.8347 & 0.5213 & 0.6027 & \textcolor{gray}{0.0455} & 0.7824 & \textbf{0.1112} \\
\bottomrule
\end{tabular}
}
\end{table}

\begin{table}[t]
\centering
\caption{Ablation study on the effect of pretraining under UPAO, measured by VRMSE ($\downarrow$). We compare our pretrained NSLoRA adaptation against full-parameter training from random initialization.}
\label{tab:ablation_pretrain}
\resizebox{\textwidth}{!}{%
\begin{tabular}{l cccc cccc}
\toprule
& \multicolumn{4}{c}{\textbf{Exact-Solution Datasets}} & \multicolumn{4}{c}{\textbf{The Well}} \\
\cmidrule(lr){2-5}\cmidrule(lr){6-9}
& \shortstack{TG\\2D} & \shortstack{Wave\\2D} & \shortstack{AdvDiff\\2D} & \shortstack{Burgers\\2D}
& \shortstack{Rayleigh\\B\'{e}nard} & \shortstack{Shear\\Flow} & \shortstack{Gray\\Scott} & \shortstack{Active\\Matter} \\
\midrule
Scratch (full-param, no pretrain) & 0.3920 & 0.7903 & 0.0673 & 0.0716 & 0.2215 & 0.5839 & 0.3366 & 0.4588 \\
Ours (NSLoRA, pretrained) & \textbf{0.0060} & \textbf{0.0218} & \textbf{0.0038} & \textbf{0.0021} & \textbf{0.1812} & \textbf{0.1112} & \textbf{0.1099} & \textbf{0.1215} \\
\bottomrule
\end{tabular}%
}
\end{table}

\vspace{-0.5em}
\subsection{Ablation Studies}
\label{sec:ablations}
\vspace{-0.5em}
\textbf{Effect of pretraining.}
We demonstrate the significance of pretraining in Table~\ref{tab:ablation_pretrain}. Optimizing all parameters from random initialization under UPAO degrades VRMSE on all eight 2D benchmarks. The degradation reaches more than an order of magnitude on the four exact-solution datasets and remains a factor of 1.2 to 5.3 on the four datasets in The Well. The pretrained weights thus provide the transferable representations that UPAO leverages, whereas their absence fails to adapt the PDE solver from random weights to good performance.

\textbf{Effect of NSLoRA.}
\label{sec:nslora_ablation}
Table~\ref{tab:ablation_orthlora} compares NSLoRA against standard LoRA~\cite{hu2022lora} on the same backbone, with both adapters initialized from a shared LoRA warm-up and only the forward path differing. NSLoRA reduces VRMSE on all eight 2D benchmarks by an average of approximately $4\%$. The complete per-channel results provided in Table~\ref{tab:app_per_channel} further reveals that these improvements are concentrated on less dominant physical quantities such as Burgers $V_x$, Shear Flow tracer, and TG pressure. This improvement directly supports the physical rebalancing motivation of NSLoRA.

To diagnose the mechanism, Table~\ref{tab:spectral} reports the effective rank of $\Delta W$ across $108$ adapter layers per checkpoint. Despite a nominal rank budget of $r\!=\!16$, the stable rank of standard LoRA lies between $1.31$ and $2.20$ on the eight 2D datasets, confirming the rank deficiency that NSLoRA addresses. NSLoRA consistently lifts the stable rank, participation ratio, and entropy rank, raising the stable rank by $5.2\%$ on average and by up to $15.6\%$ on Rayleigh--B\'enard. This provides direct empirical evidence that the orthogonalized forward path mitigates rank collapse, while the absolute stable rank remains far below the nominal budget under both forward paths.

\begin{table}[t]
\centering
\caption{Ablation study comparing NSLoRA against standard LoRA~\cite{hu2022lora}, measured by VRMSE~($\downarrow$). Both methods employ an identical rank budget and share the exact same warmed-up initialization matrices, differing exclusively in how the forward pass processes low-rank matrices.}
\label{tab:ablation_orthlora}
\footnotesize
\setlength{\tabcolsep}{5pt}
\makebox[\textwidth][c]{%
\begin{tabular*}{\textwidth}{@{\extracolsep{\fill}}l cccc cccc}
\toprule
& \multicolumn{4}{c}{\textbf{Exact-Solution Datasets}} & \multicolumn{4}{c}{\textbf{The Well}} \\
\cmidrule(lr){2-5}\cmidrule(lr){6-9}
& \shortstack{TG\\2D} & \shortstack{Wave\\2D} & \shortstack{AdvDiff\\2D} & \shortstack{Burgers\\2D}
& \shortstack{Gray\\Scott} & \shortstack{Active\\Matter} & \shortstack{Rayleigh\\B\'{e}nard} & \shortstack{Shear\\Flow} \\
\midrule
Standard LoRA~\cite{hu2022lora} & 0.0062 & 0.0224 & 0.0042 & 0.0023 & 0.1101 & 0.1249 & 0.1870 & 0.1141 \\
NSLoRA & \textbf{0.0060} & \textbf{0.0218} & \textbf{0.0038} & \textbf{0.0021} & \textbf{0.1099} & \textbf{0.1215} & \textbf{0.1812} & \textbf{0.1112} \\
\bottomrule
\end{tabular*}}
\end{table}

\begin{table}[t]
\centering
\caption{Spectral analysis of $\Delta W$ across $108$ adapter layers per checkpoint, with nominal rank budget $r\!=\!16$. Higher values indicate broader use of the rank budget.}
\label{tab:spectral}
\renewcommand{\arraystretch}{0.9}
{\footnotesize
\begin{tabular*}{\textwidth}{@{\extracolsep{\fill}} l cc cc cc @{}}
\toprule
& \multicolumn{2}{c}{Stable Rank ($\uparrow$)} & \multicolumn{2}{c}{Participation Ratio ($\uparrow$)} & \multicolumn{2}{c}{Entropy Rank ($\uparrow$)} \\
\cmidrule(lr){2-3}\cmidrule(lr){4-5}\cmidrule(lr){6-7}
Dataset & Std. LoRA & NSLoRA & Std. LoRA & NSLoRA & Std. LoRA & NSLoRA \\
\midrule
TG 2D            & 1.470 & \textbf{1.591} & 0.334 & \textbf{0.377} & 7.142 & \textbf{7.731} \\
Burgers 2D       & 1.522 & \textbf{1.557} & 0.315 & \textbf{0.346} & 6.785 & \textbf{7.267} \\
AdvDiff 2D       & 1.373 & \textbf{1.448} & 0.303 & \textbf{0.339} & 6.726 & \textbf{7.275} \\
Wave 2D          & 1.308 & \textbf{1.378} & 0.289 & \textbf{0.331} & 6.522 & \textbf{7.122} \\
Gray--Scott      & 2.198 & \textbf{2.225} & 0.406 & \textbf{0.423} & 7.930 & \textbf{8.156} \\
Active Matter    & 1.495 & \textbf{1.507} & 0.308 & \textbf{0.340} & 6.686 & \textbf{7.233} \\
Rayleigh--B\'enard& 1.367 & \textbf{1.580} & 0.287 & \textbf{0.337} & 6.412 & \textbf{7.229} \\
Shear Flow       & 1.352 & \textbf{1.393} & 0.287 & \textbf{0.314} & 6.380 & \textbf{6.864} \\
\bottomrule
\end{tabular*}
}
\end{table}

\textbf{Effect of PDE term normalization.}
Table~\ref{tab:app_norm} compares three normalization strategies for the PDE residual term. The residual scales can span multiple orders of magnitude across sub-equations, and the per-sub-equation division by $s_k^2$ used in \textsc{norm} prevents any single sub-equation from dominating the gradient signal. \textsc{rescaled\_norm} retains this division and additionally sets $\lambda_{\mathrm{BC}} = 10^4/\min_k s_k^2$, so that the boundary term keeps a comparable relative weight once the residuals have been scaled down. \textsc{rescaled\_norm} achieves superior performance on most of the eight datasets.

\begin{table}[!t]
\centering
\caption{Ablation study on the normalization strategy of the PDE residual term, measured by VRMSE ($\downarrow$). The three configurations differ only in whether $\mathcal{L}_{\mathrm{PDE}}$ divides each sub-equation residual by its own $s_k^2$, and in how $\lambda_{\mathrm{BC}}$ is set. \textsc{norm} applies the per-sub-equation normalization of Eq.~\ref{eq:pde_loss} exactly as written, with $\lambda_{\mathrm{BC}} = 10^4$. \textsc{rescaled\_norm} applies the same normalization and additionally sets $\lambda_{\mathrm{BC}} = 10^4/\min_k s_k^2$, restoring the relative weight of the boundary term after the residuals have been scaled down. \textsc{no\_norm} omits the division by $s_k^2$ in $\mathcal{L}_{\mathrm{PDE}}$ and keeps $\lambda_{\mathrm{BC}} = 10^4$.}
\label{tab:app_norm}
\resizebox{\textwidth}{!}{%
\begin{tabular}{lcccccccc}
\toprule
& \multicolumn{4}{c}{\textbf{Exact-Solution Datasets}} & \multicolumn{4}{c}{\textbf{The Well}} \\
\cmidrule(lr){2-5}\cmidrule(lr){6-9}
Strategy & \shortstack{TG\\ 2D} & \shortstack{Wave\\2D} & \shortstack{AdvDiff\\2D} & \shortstack{Burgers\\2D} & \shortstack{Gray\\Scott} & \shortstack{Active\\Matter} & \shortstack{Rayleigh\\B\'{e}nard} & \shortstack{Shear\\Flow} \\
\midrule
\textsc{rescaled\_norm} & \textbf{0.0070} & \textbf{0.0275} & \textbf{0.0049} & \textbf{0.0034} & 0.1117 & 0.0768 & 0.1884 & \textbf{0.1151} \\
\textsc{norm} & 0.0294 & 0.1007 & 0.0133 & 0.0109 & \textbf{0.1110} & 0.0767 & 0.1883 & 0.5007 \\
\textsc{no\_norm} & 0.0875 & 0.0478 & 0.0102 & 0.1303 & 0.1112 & 0.0769 & 0.1940 & 0.1341 \\
\bottomrule
\end{tabular}%
}
\end{table}

Structural ablation regarding LoRA versus full-parameter finetuning (Table~\ref{tab:app_lora_vs_full}) is reported in Appendix~\ref{app:ablations}. We further provide analytical experiments including a sensitivity analysis of the boundary weight $\lambda_{\mathrm{BC}}$ (Table~\ref{tab:app_lambda_bc}), comprehensive cross-target transfer panels for CNextU-Net and TFNO (Table~\ref{tab:app_cross_domain_full}), and an execution speed comparison between Newton-Schulz and SVD orthogonalization (Table~\ref{tab:app_ns_layer}).

\vspace{-0.75em}
\section{Conclusion}
\vspace{-0.5em}
We propose an unsupervised PDE-based finetuning framework that adapts a pretrained PDE foundation model to unseen equations strictly through the governing PDEs and boundary conditions, eliminating the requirement for interior ground-truth fields. We design a NA Transformer that accommodates varying spatial resolutions. We introduce NSLoRA to rebalance learning across heterogeneous physical quantities. Across eleven downstream datasets, our framework reduces VRMSE by a geometric-mean factor of $9.9$ relative to PDE-Transformer~\cite{holzschuh2025pdetransformer} under UPAO and remains within a factor of $2.5$ of supervised LoRA finetuning on the same backbone on seven of the eight 2D benchmarks. We outperform at least one competitive neural operator baseline on nine of the eleven datasets, while requiring no interior labels. We demonstrate that the gain from UPAO is backbone-dependent, remaining consistent on TFNO, marginal or negative on CNextU-Net, and largest and most uniform on our pretrained backbone.

\textbf{Limitations.}
The framework requires that the governing equations of the target PDE are known and that boundary observations are available throughout the trajectory, restricting applicability when such priors cannot be obtained. In addition, the PDE residual term provides weaker supervision when the simulation data carries discretization error. Evaluation is restricted to single-step prediction, leaving autoregressive stability untested.


\bibliographystyle{plainnat}
\bibliography{references}


\appendix
\newpage

\section{Architecture and Implementation Details}
\label{app:method_details}

\subsection{Architecture Hyperparameters}
\label{app:arch}

Table~\ref{tab:app_arch} lists the concrete hyperparameter values for the backbone in Section~\ref{sec:architecture} and the LoRA adapter in Section~\ref{sec:orthlora}. The first block lists the backbone configuration shared across pretraining and finetuning, and the second block lists the adapter configuration applied during the adaptation stage.

\begin{table}[ht]
\centering
\caption{Hyperparameters of the NA Transformer backbone and the LoRA adapter.}
\label{tab:app_arch}
\begin{tabular}{ll}
\toprule
Hyperparameter & Value \\
\midrule
History window $T_{\mathrm{in}}$ & 8 \\
Channel count $C$ & 18 (3 velocity components, 15 scalar slots) \\
Patch size & $16 \times 16$ \\
Post-patch feature map & $4 \times 4$ \\
Encoder strided convs & 2 \\
Embedding dimension $H$ & 768 \\
Transformer layers $L$ & 12 \\
Attention heads & 12 \\
NA window size $k$ & $5 \times 5$ \\
Post-smoothing & three $7 \times 7$ convolutions \\
Total parameters & $\sim$182M \\
\midrule
LoRA rank $r$ & 16 \\
LoRA scaling $\alpha$ & 32 \\
\bottomrule
\end{tabular}
\end{table}

\subsection{Aspect-Ratio NA Variants and Routing}
\label{app:aspect_na}

The downstream datasets cover three regimes of spatial aspect ratio: square ($1{:}1$, e.g.\ Active Matter $256 \times 256$ and Gray--Scott $128 \times 128$), moderately elongated ($1{:}2$, e.g.\ Shear Flow $512 \times 256$), and strongly elongated ($1{:}4$, e.g.\ Rayleigh--B\'enard $512 \times 128$). A single neighborhood attention kernel of fixed shape would either truncate the long axis below the natural correlation length or oversample the short axis at quadratic cost. We therefore instantiate three 2D variants with geometrically congruent spatial kernels, summarised in Table~\ref{tab:app_na_variants}. Each batch is routed by its post-patch token aspect ratio to the matching variant, and only that variant receives gradient updates from the batch. Variants share neither weights nor learning rate schedules with each other.

\begin{table}[ht]
\centering
\caption{NA variants for 2D inputs. Aspect ratio is computed on the post-patch token grid (input resolution divided by the patch side $p = 16$). Each variant is instantiated as an independent module and exclusively updated by batches that fall in its aspect-ratio range.}
\label{tab:app_na_variants}
\begin{tabular}{lccl}
\toprule
Variant & Spatial kernel & Aspect-ratio range & Example dataset \\
\midrule
$\mathrm{NA}^{(2)}_{1:1}$ & $5 \times 5$ & $[1, 1.5)$ & Active Matter, Gray--Scott \\
$\mathrm{NA}^{(2)}_{1:2}$ & $5 \times 9$ & $[1.5, 3.0)$ & Shear Flow \\
$\mathrm{NA}^{(2)}_{1:4}$ & $5 \times 17$ & $[3.0, \infty)$ & Rayleigh--B\'enard \\
\bottomrule
\end{tabular}
\end{table}

The $1{:}1$ variant uses the base kernel $k = 5$ in Table~\ref{tab:app_arch}. The $1{:}2$ and $1{:}4$ variants stretch the kernel along the elongated axis to $9$ and $17$ tokens respectively, preserving a comparable geometric coverage of the underlying physical domain. Since virtually all current 3D datasets are cubic, we only train the $1{:}1{:}1$ variant for 3D.

\subsection{Channel Slot Mapping}
\label{app:slot_mapping}

The model operates on a unified $C = 18$ channel representation, partitioned into $3$ vector slots $(V_x, V_y, V_z)$ and $15$ scalar slots fixed across all datasets. Each downstream dataset writes physical fields into the slots that match its semantics and leaves the remaining slots zero-filled in the input tensor. A binary channel mask of length $18$ marks the active slots, and the boundary condition term $\mathcal{L}_{\mathrm{BC}}$, the PDE residual term $\mathcal{L}_{\mathrm{PDE}}$, and the reported VRMSE are evaluated exclusively on the active channels. The fixed slot inventory is listed in Table~\ref{tab:app_scalar_slots}.

\begin{table}[ht]
\centering
\caption{Scalar channel slots in the unified $18$-channel representation. Each downstream dataset writes its physical fields into the slots whose semantics match, and the remaining slots are zero-filled in the input tensor. The three vector slots $(V_x, V_y, V_z)$ are listed separately and not enumerated in this table.}
\label{tab:app_scalar_slots}
\begin{tabular}{cll}
\toprule
Index & Slot name & Physical interpretation \\
\midrule
0  & buoyancy                 & buoyancy or external forcing \\
1  & concentration\_rho       & single-species concentration \\
2  & concentration\_u         & reaction-diffusion activator \\
3  & concentration\_v         & reaction-diffusion inhibitor \\
4  & density                  & fluid or material density $\rho$ \\
5  & electron\_fraction       & electron fraction \\
6  & energy                   & internal energy \\
7  & entropy                  & entropy \\
8  & geometry                 & geometry mask such as airfoil shape \\
9  & gravitational\_potential & gravitational potential \\
10 & height                   & water height for shallow water \\
11 & passive\_tracer          & passive tracer or particle field \\
12 & pressure                 & pressure $p$ \\
13 & speed\_of\_sound         & material speed of sound \\
14 & temperature              & temperature $T$ \\
\bottomrule
\end{tabular}
\end{table}

\subsection{NSLoRA Implementation Details}
\label{app:nslora_impl}

NSLoRA orthogonalises each low-rank matrix via Newton-Schulz iteration~\cite{jordan2024muon} before composing the rank-$r$ update. Each iteration applies a quintic polynomial to the Gram product of the current iterate. With coefficients $(a, b, c) = (3.4445, -4.7750, 2.0315)$ and five iterations, the iterate converges to a matrix whose nontrivial singular values lie within a narrow band around one~\cite{jordan2024muon}. We denote the operator by $\mathrm{NS}_n(\cdot)$.

For an iterate $X$, one Newton-Schulz step is given by
\begin{equation}
X \leftarrow a\, X + b\, X X^{\top} X + c\, (X X^{\top})^2 X.
\label{eq:ns_step}
\end{equation}
The full orthogonalisation $\mathrm{NS}_n(M)$ initialises $X \leftarrow M / (\|M\|_F + \epsilon)$, repeats Eq.~\ref{eq:ns_step} for $n$ iterations, and returns the final iterate.

The NSLoRA forward pass operates on a single LoRA layer with pretrained weight $W_0 \in \mathbb{R}^{d_{\mathrm{out}} \times d_{\mathrm{in}}}$ and trainable factors $A \in \mathbb{R}^{r \times d_{\mathrm{in}}}, B \in \mathbb{R}^{d_{\mathrm{out}} \times r}$. We first compute the Frobenius norms $\nu_A = \|A\|_F$ and $\nu_B = \|B\|_F$. We then orthogonalize both factors as $\hat{A} = \mathrm{NS}_5(A)$ and $\hat{B} = \mathrm{NS}_5(B)$, form the rank-$r$ update $\Delta W = \nu_A \nu_B \hat{B} \hat{A}$, and return the effective weight $W_0 + \Delta W$. The two scalar magnitudes $\nu_A$ and $\nu_B$ remain trainable through the standard LoRA gradient path on $A, B$, while the orthogonalized matrices $\hat{A}, \hat{B}$ supply the directional content of the update.

The NSLoRA forward path is preceded by a standard LoRA warm-up that jointly optimizes $A$ and $B$ under the unsupervised PDE-based objective with the conventional forward $\Delta W = (\alpha/r)\, BA$. This warm-up brings the product $BA$ into a task-relevant subspace before any orthogonalization is applied. We switch from the warm-up to the NSLoRA forward path once the validation VRMSE fails to improve for five consecutive epochs, while retaining the warm-up matrices $A, B$ and the AdamW optimizer state across the switch. Training then continues under the same objective and optimizer schedule, with the forward path being the only change.

\noindent \textbf{Comparative Setup and Fairness.} 
To evaluate the efficacy of the NSLoRA forward path, we conduct a controlled comparison against the standard LoRA baseline. For a fair comparison, both methods share an identical training infrastructure, including hyperparameter configurations, UPAO, and the AdamW optimizer state inherited from the warm-up phase. The only distinction resides in the forward pass calculation: while the baseline maintains the conventional $\Delta W = (\alpha/r) BA$ throughout, NSLoRA transitions to the orthogonalized update via Eq.~\ref{eq:ns_step} after the warm-up. Both variants are trained until convergence, determined by either the early stopping criterion (five consecutive epochs without VRMSE improvement) or reaching the maximum limit of 30 epochs. All quantitative metrics are systematically reported in Table~\ref{tab:app_per_channel}.

\subsection{PDE residual Implementation}
\label{app:PDEresidual}
\textbf{Upwind Discretization for Convective Nonlinearities.}
For convective terms such as $\nabla \cdot (\mathbf{v} u)$, the residual term uses a conservative upwind scheme~\cite{chiu2022can} with second-order face-value interpolation. Let $f_i$ denote the cell-centered flux at grid point $i$. The two candidate face values at $i + 1/2$ are
\begin{equation}
\hat{f}_{i + 1/2}^{\,+} = \tfrac{3}{2} f_i - \tfrac{1}{2} f_{i-1}, \qquad
\hat{f}_{i + 1/2}^{\,-} = \tfrac{3}{2} f_{i+1} - \tfrac{1}{2} f_{i+2},
\label{eq:upwind}
\end{equation}
where $+$ and $-$ denote the left- and right-biased stencils. The final face value is selected from the two candidates based on the sign of the face velocity. The same construction applies symmetrically to the face at $i - 1/2$.

\textbf{Ghost-Cell Extrapolation for Non-Periodic Domains.}
For non-periodic boundaries, central-difference stencils at the outermost interior points require neighbours that lie outside $\partial\Omega$. We populate a single virtual layer of grid points (ghost cells) immediately adjacent to $\partial\Omega$ by second-order one-sided polynomial extrapolation of the three nearest interior values, restoring a uniform stencil width across the entire grid. The boundary band $\mathcal{S}_b$ itself is excluded from the residual evaluation, so ghost cells exclusively support spatial derivatives of operators internal to $\mathcal{N}$ and never enter $\mathcal{L}_{\mathrm{BC}}$.

\section{Datasets}
\label{app:datasets}

We train the framework on a single pretraining corpus and evaluate it across eleven downstream datasets. The evaluation suite includes seven custom-synthesized datasets with closed-form analytical solutions (Section~\ref{app:exact}) alongside four high-fidelity simulation datasets from The Well benchmark~\cite{ohana2024thewell} (Section~\ref{app:wellds}). The pretraining corpus consists of a six-subset selection from PDEBench~\cite{takamoto2022pdebench} (Section~\ref{app:pretrain}). Table~\ref{tab:dataset_overview} provides a complete overview of these data configurations. Seven of the eleven downstream datasets correspond to physical equations completely unseen during the pretraining phase.

\begin{table}[ht]
\centering
\caption{Overview of pretraining and downstream datasets. Resolution is reported per spatial axis. $T$ denotes the number of stored time steps per trajectory. ``Unseen'' marks downstream datasets whose governing PDE is not present in the pretraining corpus.}
\label{tab:dataset_overview}
\resizebox{\textwidth}{!}{%
\begin{tabular}{llccccc}
\toprule
Dataset & PDE & Dim & Resolution & $T$ & \# Trajectories & Unseen \\
\midrule
\multicolumn{7}{l}{\textit{Pretraining (PDEBench~\cite{takamoto2022pdebench})}} \\
\midrule
1D CFD & Compressible Navier-Stokes & 1D & 1024 & 21 & 10{,}000 & --- \\
2D CFD & Compressible Navier-Stokes & 2D & $128{\times}128$ & 21 & 4{,}000 & --- \\
3D CFD & Compressible Navier-Stokes & 3D & $64{\times}64{\times}64$ & 21 & 100 & --- \\
Shallow Water & Shallow-water equations & 2D & $128{\times}128$ & 101 & 1{,}000 & --- \\
Diffusion-Reaction & Two-species reaction-diffusion & 2D & $128{\times}128$ & 101 & 1{,}000 & --- \\
Incompressible NS & Incompressible Navier-Stokes & 2D & $512{\times}512$ & 21 & 1{,}000 & --- \\
\midrule
\multicolumn{7}{l}{\textit{Exact-solution finetuning datasets (Section~\ref{app:exact})}} \\
\midrule
2D TG & Incompressible Navier-Stokes & 2D & $256{\times}256$ & 101 & 100 & no \\
2D Wave & Linear wave equation & 2D & $256{\times}256$ & 101 & 100 & yes \\
2D Advection--Diffusion & Linear advection--diffusion & 2D & $256{\times}256$ & 101 & 150 & yes \\
2D Burgers & Viscous Burgers (Cole--Hopf) & 2D & $256{\times}256$ & 101 & 200 & yes \\
1D Burgers & Viscous Burgers (Cole--Hopf) & 1D & 1024 & 101 & 1{,}000 & yes \\
1D Advection & Linear advection & 1D & 1024 & 101 & 1{,}000 & yes \\
3D Advection & Linear advection & 3D & $64^3$ & 21 & 50 & yes \\
\midrule
\multicolumn{7}{l}{\textit{The Well downstream datasets (Section~\ref{app:wellds})}} \\
\midrule
Active Matter & Smoluchowski + Stokes flow & 2D & $256{\times}256$ & 81 & 175 & yes \\
Rayleigh--B\'enard & Boussinesq + thermal & 2D & $512{\times}128$ & 200 & 100 & partial \\
Shear Flow & Incompressible Navier-Stokes & 2D & $512{\times}256$ & 200 & 50 & no \\
Gray--Scott & Two-species reaction-diffusion & 2D & $128{\times}128$ & 1{,}001 & 200 & no \\
\bottomrule
\end{tabular}}
\end{table}

\subsection{Pretraining Dataset (PDEBench)}
\label{app:pretrain}

We pretrain on six subsets of PDEBench~\cite{takamoto2022pdebench}. For each subset we write the governing equation explicitly. Parameter ranges, initial-condition samplers, and solver implementations follow the original benchmark release and we refer the reader to~\cite{takamoto2022pdebench} for those details.

\subsubsection{1D and 2D Compressible Navier-Stokes (1D CFD, 2D CFD)}
The density $\rho$, velocity $\mathbf{u}$, and pressure $p$ satisfy
\begin{align}
\partial_t \rho + \nabla \cdot (\rho \mathbf{u}) &= 0, \\
\rho(\partial_t \mathbf{u} + \mathbf{u} \cdot \nabla \mathbf{u}) &= -\nabla p + \eta \Delta \mathbf{u} + (\zeta + \eta/3)\nabla(\nabla \cdot \mathbf{u}), \\
\partial_t\!\left(\tfrac{3}{2} p + \tfrac{1}{2}\rho \|\mathbf{u}\|^2\right) &= -\nabla \cdot \bigl[(\varepsilon + p + \tfrac{1}{2}\rho\|\mathbf{u}\|^2) \mathbf{u} - \mathbf{u} \cdot \boldsymbol{\sigma}'\bigr],
\end{align}
where $\eta$ and $\zeta$ are shear and bulk viscosities and $\boldsymbol{\sigma}'$ is the viscous stress tensor. We use all three PDEBench 1D CFD configurations and the three random-periodic 2D CFD configurations. Data are downsampled from their native PDEBench resolutions to $1024$ (1D) and $128{\times}128$ (2D). The first 21 time steps of each trajectory are retained.

\textbf{Parameter distributions.}
Both the 1D and 2D subsets sweep shear viscosity $\eta$ and bulk viscosity $\zeta$ across the three configurations released by PDEBench, with random initial density, velocity, and pressure profiles drawn under periodic boundary conditions. Full parameter values and initial-condition distributions are listed in Appendix~D of~\cite{takamoto2022pdebench}.

\subsubsection{3D Compressible Navier-Stokes (3D CFD)}
The governing equations are the three-dimensional version of the system above. We include the PDEBench 3D CFD subset at resolution $64{\times}64{\times}64$ with $T=21$ time steps, which is the only 3D subset in our pretraining corpus.

\textbf{Parameter distributions.}
Trajectories vary in random initial density, velocity, and pressure profiles sampled by the PDEBench release, with shear and bulk viscosities fixed at the values reported in Appendix~D of~\cite{takamoto2022pdebench}.

\subsubsection{Shallow Water Equations}
Writing $h(\mathbf{x}, t)$ for water depth and $\mathbf{u}$ for depth-averaged velocity, the PDEBench radial dam-break scenario is governed by
\begin{align}
\partial_t h + \nabla \cdot (h \mathbf{u}) &= 0, \\
\partial_t (h \mathbf{u}) + \nabla \cdot \bigl(h \mathbf{u} \otimes \mathbf{u} + \tfrac{1}{2} g_r h^2 \mathbb{I}\bigr) &= - g_r h \nabla b,
\end{align}
with gravitational acceleration $g_r$ and bathymetry $b(\mathbf{x})$. We use the 2D Shallow Water subset at $128{\times}128$ with $T=101$ snapshots per trajectory.

\textbf{Parameter distributions.}
Trajectories vary in the radial dam-break initial condition, with the inner-pool height, outer-pool height, and pool radius drawn per trajectory from the distributions defined in Appendix~D of~\cite{takamoto2022pdebench}.

\subsubsection{Diffusion--Reaction (2D)}
Two chemical species $u, v$ evolve under FitzHugh--Nagumo-type kinetics,
\begin{align}
\partial_t u &= D_u \Delta u + u - u^3 - k - v, \\
\partial_t v &= D_v \Delta v + u - v,
\end{align}
on the domain $[-1, 1]^2$ with Neumann boundary conditions, generated via a stochastic finite-volume scheme. We use the full PDEBench Diffusion-Reaction subset at $128{\times}128$ with $T=101$.

\textbf{Parameter distributions.}
Diffusion coefficients $D_u, D_v$ and reaction parameter $k$ are fixed at the values released by PDEBench. Trajectories vary in their stochastic initial conditions sampled per trajectory following Appendix~D of~\cite{takamoto2022pdebench}.

\subsubsection{Incompressible Navier-Stokes (2D)}
With velocity $\mathbf{u}$ and pressure $p$,
\begin{align}
\partial_t \mathbf{u} + \mathbf{u} \cdot \nabla \mathbf{u} &= - \nabla p + \nu \Delta \mathbf{u} + \mathbf{f}, \\
\nabla \cdot \mathbf{u} &= 0.
\end{align}
We use the 2D incompressible subset at native $512{\times}512$, downsampled to $512{\times}512$ (no change) and kept at $T=21$.

\textbf{Parameter distributions.}
Trajectories vary in random forcing $\mathbf{f}$ and random initial velocity sampled per trajectory by the PDEBench release. Viscosity $\nu$ follows the value reported in Appendix~D of~\cite{takamoto2022pdebench}.

\subsection{Exact-Solution Finetuning Datasets}
\label{app:exact}

We generate seven datasets for which the solution admits a closed form in terms of the initial condition. All datasets are stored on periodic grids with boundary-exclusive discretization ($x_i = i \cdot L / N$ for $i=0,\ldots,N-1$) so that periodic differentiation via the fast Fourier transform is exact. Analytical ground truth eliminates discretization error as a confounder for the residual loss evaluation.

\subsubsection{2D Taylor--Green}

\textbf{Governing equations.}
The 2D Taylor--Green is a classical test case for the incompressible Navier-Stokes equations and admits a closed-form decaying solution on the doubly periodic domain $[0, 2\pi]^2$:
\begin{align}
\partial_t \mathbf{u} + (\mathbf{u} \cdot \nabla) \mathbf{u} &= -\nabla p + \nu \Delta \mathbf{u}, \qquad \nabla \cdot \mathbf{u} = 0,
\end{align}
with the analytical solution
\begin{align}
u(x, y, t) &= -\cos(x)\sin(y)\, e^{-2\nu t}, \\
v(x, y, t) &= \phantom{-}\sin(x)\cos(y)\, e^{-2\nu t}, \\
p(x, y, t) &= -\tfrac{1}{4}\bigl(\cos(2x) + \cos(2y)\bigr) e^{-4\nu t}.
\end{align}
Velocity magnitudes decay at rate $2\nu$ and pressure at rate $4\nu$.

\textbf{Spatiotemporal domain.}
Fields are evaluated analytically at each grid point and time step on a $256 \times 256$ uniform grid with $\Delta x = 2\pi/256$, over the time interval $t \in [0, 1]$ at $\Delta t = 0.01$, yielding $T = 101$ stored snapshots per trajectory.

\textbf{Parameter distributions.}
$\nu \sim \mathrm{Uniform}(0.01, 0.1)$ sampled independently per trajectory.

\textbf{Fields available in the data.}
velocity $\mathbf{u} = (u, v)$ (vector field), pressure $p$ (scalar field).

\textbf{References.}
Taylor and Green~\cite{taylor1937mechanism} for the original derivation.

\subsubsection{2D Wave Equation}

\textbf{Governing equations.}
The second-order linear wave equation on $[0, 2\pi)^2$ with wave speed $c$ reads
\begin{align}
\partial_{tt} u = c^2 \Delta u.
\end{align}
We express it as a first-order system by introducing $w = \partial_t u$,
\begin{align}
\partial_t u = w, \qquad \partial_t w = c^2 \Delta u.
\end{align}
For a multi-mode Fourier initial condition, the exact solution is a superposition of standing waves,
\begin{align}
u(x, y, t) &= \sum_{(m, n)} A_{m n} \cos(m x + n y + \phi_{m n}) \cos(\omega_{m n} t), \\
w(x, y, t) &= -\sum_{(m, n)} A_{m n}\, \omega_{m n} \cos(m x + n y + \phi_{m n}) \sin(\omega_{m n} t),
\end{align}
with $\omega_{m n} = c\sqrt{m^2 + n^2}$.

\textbf{Spatiotemporal domain.}
Fields are evaluated analytically on a $256 \times 256$ periodic grid over the time interval $t \in [0, 1]$ at $\Delta t = 0.01$, yielding $T = 101$ stored snapshots per trajectory. Each trajectory uses a random subset of modes with $|m|, |n| \leq 3$ and amplitudes, phases drawn from uniform distributions.

\textbf{Parameter distributions.}
$c \sim \mathrm{Uniform}(1.0, 3.0)$ sampled independently per trajectory. Each trajectory uses every mode $(m, n)$ with $|m|, |n| \leq 3$ and $(m, n) \neq (0, 0)$, with raw amplitudes $A_{m n}^{\mathrm{raw}} \sim \mathrm{Uniform}(0.1, 1.0)$ scaled by $1/\sqrt{m^2 + n^2}$ for spectral decay, and phases $\phi_{m n} \sim \mathrm{Uniform}(0, 2\pi)$.

\textbf{Fields available in the data.}
displacement $u$ (scalar field), time derivative $w = \partial_t u$ (scalar field).

\textbf{References.}
The standing-wave Fourier-series solution follows from separation of variables on the periodic torus. The classical derivation of the wave equation and its eigenfunction expansion are detailed in Section 2.4 of the textbook by Evans~\cite{evans2010partial}.

\subsubsection{2D Advection--Diffusion}

\textbf{Governing equations.}
A scalar concentration $u$ is transported by a constant velocity $(a, b)$ and simultaneously diffuses,
\begin{align}
\partial_t u + a \, \partial_x u + b \, \partial_y u = \nu \Delta u.
\end{align}
On $[0, 2\pi]^2$ with periodic boundary conditions, each Fourier mode translates with the velocity while its amplitude decays,
\begin{align}
u(x, y, t) = \sum_{(m, n)} A_{m n}\, e^{-\nu (m^2 + n^2) t} \cos\!\bigl(m (x - a t) + n (y - b t) + \phi_{m n}\bigr).
\end{align}

\textbf{Spatiotemporal domain.}
Fields are evaluated analytically on a $256 \times 256$ periodic grid over the time interval $t \in [0, 1]$ at $\Delta t = 0.01$, yielding $T = 101$ stored snapshots per trajectory. Modes satisfy $|m|, |n| \leq 3$, excluding $(0, 0)$.

\textbf{Parameter distributions.}
$\nu \sim \mathrm{Uniform}(0.01, 0.1)$ sampled independently per trajectory, with fixed advection velocity $(a, b) = (1.0, 0.5)$. Per trajectory, $N_{\mathrm{modes}} \in \{8, \ldots, 15\}$ modes are drawn without replacement from $\{(m, n) : |m|, |n| \leq 3\} \setminus \{(0, 0)\}$, with amplitudes $A_{m n} \sim \mathrm{Uniform}(0.1, 1.0)$ and phases $\phi_{m n} \sim \mathrm{Uniform}(0, 2\pi)$.

\textbf{Fields available in the data.}
concentration $u$ (scalar field).

\textbf{References.}
The closed-form Fourier-mode solution follows from separation of variables and the diagonal action of constant-coefficient advection-diffusion on Fourier modes. See Evans~\cite{evans2010partial}, Section 2.5 for the heat-kernel construction that underlies this expansion.

\subsubsection{2D Burgers}

\textbf{Governing equations.}
The viscous Burgers system in two dimensions reads
\begin{align}
\partial_t u + u\, \partial_x u + v\, \partial_y u &= \nu \Delta u, \\
\partial_t v + u\, \partial_x v + v\, \partial_y v &= \nu \Delta v.
\end{align}
For irrotational flow ($\partial_x v = \partial_y u$), the Cole--Hopf transform $\mathbf{u} = -2 \nu \nabla (\ln \theta)$ linearises the system into the heat equation $\partial_t \theta = \nu \Delta \theta$. Choosing $\theta$ as a separable product,
\begin{align}
\theta(x, y, t) = f(x, t) \cdot g(y, t), \qquad f = 1 + \epsilon_x e^{-\nu k_x^2 t} \cos(k_x x), \qquad g = 1 + \epsilon_y e^{-\nu k_y^2 t} \cos(k_y y),
\end{align}
yields the closed-form solution
\begin{align}
u = -2 \nu \, \partial_x \ln f, \qquad v = -2 \nu \, \partial_y \ln g.
\end{align}

\textbf{Spatiotemporal domain.}
Fields are evaluated analytically on a $256 \times 256$ periodic grid over the time interval $t \in [0, 1]$ at $\Delta t = 0.01$, yielding $T = 101$ stored snapshots per trajectory.

\textbf{Parameter distributions.}
$\nu \sim \mathrm{Uniform}(0.01, 0.1)$, $\epsilon_x, \epsilon_y \sim \mathrm{Uniform}(0.1, 0.5)$, $k_x, k_y \in \{1, 2, 3\}$. The cross product of $(k_x, k_y)$ with six random samples per combination yields 200 trajectories.

\textbf{Fields available in the data.}
velocity $\mathbf{u} = (u, v)$ (vector field).

\textbf{References.}
Cole~\cite{cole1951quasi} and Hopf~\cite{hopf1950partial} introduced the linearising substitution $\mathbf{u} = -2\nu\nabla(\ln\theta)$ that maps Burgers to the heat equation. The separable closed-form solution we use is among those tabulated in Benton and Platzman~\cite{benton1972table}.

\subsubsection{1D Burgers}

\textbf{Governing equations.}
The one-dimensional viscous Burgers equation is
\begin{align}
\partial_t u + u \, \partial_x u = \nu \, \partial_{xx} u.
\end{align}
The Cole--Hopf substitution $u = -2\nu \, \partial_x \ln \theta$ maps it to the one-dimensional heat equation $\partial_t \theta = \nu \, \partial_{xx} \theta$, whose periodic Fourier-series solution yields
\begin{align}
\theta(x, t) &= 1 + \sum_k \epsilon_k \, e^{-\nu k^2 t} \cos(k x + \phi_k), \\
u(x, t) &= \frac{2 \nu \sum_k \epsilon_k k \sin(k x + \phi_k) e^{-\nu k^2 t}}{1 + \sum_k \epsilon_k \cos(k x + \phi_k) e^{-\nu k^2 t}}.
\end{align}
The amplitudes $\epsilon_k$ are chosen small enough that $\theta$ remains strictly positive on the grid, ensuring the logarithm is well-defined throughout the trajectory.

\textbf{Spatiotemporal domain.}
Fields are evaluated analytically on a 1D periodic grid of $N = 1024$ points on $[0, 2\pi]$ over the time interval $t \in [0, 1]$ at $\Delta t = 0.01$, yielding $T = 101$ stored snapshots per trajectory.

\textbf{Parameter distributions.}
$\nu \sim \mathrm{Uniform}(0.005, 0.1)$ sampled independently per trajectory. Each initial condition uses $N_{\mathrm{modes}} \in \{3, \ldots, 8\}$ Fourier modes with wavenumbers $k \in \{1, \ldots, 5\}$, raw amplitudes $\epsilon_k \sim \mathrm{Uniform}(0.1, 0.5)$ rescaled so that $\sum_k \epsilon_k \sim \mathrm{Uniform}(0.3, 0.9)$ to keep $\theta$ strictly positive on the grid, and phases $\phi_k \sim \mathrm{Uniform}(0, 2\pi)$.

\textbf{Fields available in the data.}
velocity $u$ (vector field, aligned with $V_x$ channel).

\textbf{References.}
Cole~\cite{cole1951quasi} and Hopf~\cite{hopf1950partial} introduced the substitution $u = -2\nu\partial_x \ln\theta$ that maps the one-dimensional viscous Burgers equation to the heat equation, from which our periodic Fourier-series solution follows. Benton and Platzman~\cite{benton1972table} catalogue the analogous closed-form solutions of the one-dimensional Burgers equation.

\subsubsection{1D Advection}

\textbf{Governing equations.}
The linear advection equation on the periodic interval $[0, 2\pi]$ reads
\begin{align}
\partial_t u + a \, \partial_x u = 0,
\end{align}
with exact solution $u(x, t) = u_0(x - a t)$ by the method of characteristics. For a multi-mode sinusoidal initial condition $u_0(x) = \sum_j A_j \sin(k_j x + \phi_j)$, the translated form
\begin{align}
u(x, t) = \sum_j A_j \sin\!\bigl(k_j (x - a t) + \phi_j\bigr)
\end{align}
preserves amplitude at every time step with zero numerical error.

\textbf{Spatiotemporal domain.}
Fields are evaluated analytically on a 1D periodic grid of $N = 1024$ points over the time interval $t \in [0, 1]$ at $\Delta t = 0.01$, yielding $T = 101$ stored snapshots per trajectory.

\textbf{Parameter distributions.}
$a \sim \mathrm{Uniform}(0.5, 3.0)$ sampled independently per trajectory. Each initial condition uses $N_{\mathrm{modes}} \in \{5, \ldots, 12\}$ sinusoidal modes with wavenumbers $k_j \in \{1, \ldots, 8\}$, amplitudes $A_j \sim \mathrm{Uniform}(0.3, 1.5)$, and phases $\phi_j \sim \mathrm{Uniform}(0, 2\pi)$.

\textbf{Fields available in the data.}
scalar $u$.

\textbf{References.}
Evans~\cite{evans2010partial}, Section 2.1 derives the closed-form solution $u(x, t) = u_0(x - at)$ via the method of characteristics. The same scenario with periodic boundary conditions is provided as the \texttt{adv} benchmark in APEBench~\cite{koehler2024apebench}, Section 2.

\subsubsection{3D Advection}

\textbf{Governing equations.}
The linear advection equation in three dimensions on $[0, 2\pi]^3$ is
\begin{align}
\partial_t u + a\, \partial_x u + b\, \partial_y u + c\, \partial_z u = 0,
\end{align}
with closed-form solution $u(\mathbf{x}, t) = u_0(\mathbf{x} - \mathbf{c}\, t)$ by componentwise extension of the method of characteristics. We generate trajectories from multi-mode Fourier initial conditions,
\begin{align}
u(\mathbf{x}, t) = \sum_{(l, m, n)} A_{l m n} \cos\!\bigl(l (x - a t) + m (y - b t) + n (z - c t) + \phi_{l m n}\bigr).
\end{align}

\textbf{Spatiotemporal domain.}
Fields are evaluated analytically on a $64 \times 64 \times 64$ periodic grid over the time interval $t \in [0, 1]$ at $\Delta t = 0.05$, yielding $T = 21$ stored snapshots per trajectory.

\textbf{Parameter distributions.}
$(a, b, c) \sim \mathrm{Uniform}(0.5, 2.0)^3$ sampled independently per component. Per trajectory, $N_{\mathrm{modes}} \in \{8, \ldots, 15\}$ modes are drawn without replacement from $\{(l, m, n) : |l|, |m|, |n| \leq 2\} \setminus \{(0, 0, 0)\}$, with amplitudes $A_{l m n} \sim \mathrm{Uniform}(0.1, 1.0)$ and phases $\phi_{l m n} \sim \mathrm{Uniform}(0, 2\pi)$.

\textbf{Fields available in the data.}
scalar $u$.

\textbf{References.}
The closed-form solution $u(\mathbf{x}, t) = u_0(\mathbf{x} - \mathbf{c} t)$ follows by componentwise extension of the method of characteristics. The one-dimensional case is treated in Evans~\cite{evans2010partial}, Section 2.1. Appendix~E.1 of APEBench~\cite{koehler2024apebench} provides the same isotropic 3D advection scenario with periodic boundary conditions.

\subsection{The Well Downstream Datasets}
\label{app:wellds}

We additionally finetune on four simulation datasets from the Well benchmark~\cite{ohana2024thewell}. Because the governing equations, initial-condition distributions, and numerical solvers are described in full in Appendix~C of~\cite{ohana2024thewell}, we restrict the discussion here to the fields and trajectory subsets we use.

\subsubsection{Active Matter}
A continuum kinetic theory of rod-like active particles immersed in a Stokes fluid, governed by a coupled Smoluchowski--Stokes system with orientation-resolved distribution $\Psi$. We use the concentration scalar and the two-component velocity field from the original Well release at resolution $256 \times 256$ with $T = 81$. Other fields such as the orientation tensor and the strain-rate tensor are dropped. 

\textbf{Parameter distributions.}
The active dipole strength is swept over $\alpha \in \{-1, -2, -3, -4, -5\}$ with $\beta = 0.8$ fixed and the alignment strength swept over $\zeta \in \{1, 3, 5, 7, 9, 11, 13, 15, 17\}$, exactly as released in Appendix~C.2 of the Well~\cite{ohana2024thewell}.

\textbf{Fields available in the data.}
concentration field denoted ``conc'' (scalar field), velocity $\mathbf{u} = (V_x, V_y)$ (vector field).

\textbf{References.}
Governing equations, simulation details, and parameter sweep: Appendix~C.2 of~\cite{ohana2024thewell}.

\subsubsection{Rayleigh--B\'enard Convection}
Boussinesq-approximation incompressible Navier-Stokes coupled with a temperature (buoyancy) field on a horizontally periodic rectangular domain with no-slip vertical walls. We use the buoyancy, pressure, and two-component velocity fields at $512 \times 128$ with $T = 200$. 

\textbf{Parameter distributions.}
The Rayleigh number is swept over $\mathrm{Ra} \in \{10^6, 10^7, 10^8, 10^9, 10^{10}\}$, the Prandtl number over $\mathrm{Pr} \in \{0.1, 0.2, 0.5, 1.0, 2.0, 5.0, 10.0\}$, and the initial buoyancy perturbation over $\delta b_0 \in \{0.2, 0.4, 0.6, 0.8, 1.0\}$, exactly as released in Appendix~C.10 of the Well~\cite{ohana2024thewell}.

\textbf{Fields available in the data.}
buoyancy field denoted ``buoy'' (scalar field), pressure $p$ (scalar field), velocity $\mathbf{u} = (V_x, V_y)$ (vector field).

\textbf{References.}
Governing equations, simulation details, and parameter sweep: Appendix~C.10 of~\cite{ohana2024thewell}.

\subsubsection{Shear Flow}
Periodic incompressible Navier-Stokes in a shear-flow configuration with two fluid layers sliding past each other at different Reynolds and Schmidt numbers. We use the passive tracer, pressure, and two-component velocity fields at $512 \times 256$ with $T = 200$. 

\textbf{Parameter distributions.}
The Reynolds number is swept over $\mathrm{Re} \in \{10^4, 5 \times 10^4, 10^5, 5 \times 10^5\}$ and the Schmidt number over $\mathrm{Sc} \in \{0.1, 0.2, 0.5, 1.0, 2.0, 5.0, 10.0\}$. Initial conditions vary in $n_{\mathrm{shear}} \in \{2, 4\}$ shear layers, $n_{\mathrm{blobs}} \in \{2, 3, 4, 5\}$ tracer blobs, and shear-width factor $w \in \{0.25, 0.5, 1.0, 2.0, 4.0\}$, exactly as released in Appendix~C.12 of the Well~\cite{ohana2024thewell}.

\textbf{Fields available in the data.}
passive tracer field denoted ``tracer'' (scalar field), pressure $p$ (scalar field), velocity $\mathbf{u} = (V_x, V_y)$ (vector field).

\textbf{References.}
Governing equations, simulation details, and parameter sweep: Appendix~C.12 of~\cite{ohana2024thewell}.

\subsubsection{Gray--Scott reaction-diffusion}
Two-species reaction-diffusion with feed rate $f$ and kill rate $k$ controlling the pattern formation regime. We use the two concentration fields at $128 \times 128$ with $T = 1{,}001$. The six $(f, k)$ pattern configurations released in the Well (gliders, bubbles, maze, worms, spirals, spots) are all included.

\textbf{Parameter distributions.}
Diffusion constants are fixed at $\delta_u = 2 \times 10^{-5}$ and $\delta_v = 10^{-5}$. The feed and kill rates are fixed within each of the six configurations: gliders $(f, k) = (0.014, 0.054)$, bubbles $(0.098, 0.057)$, maze $(0.029, 0.057)$, worms $(0.058, 0.065)$, spirals $(0.018, 0.051)$, and spots $(0.030, 0.062)$. Trajectories within each configuration vary in their initial conditions sampled following Appendix~C.5 of~\cite{ohana2024thewell}.

\textbf{Fields available in the data.}
concentrations of the two reaction-diffusion species $A$ and $B$ (scalar fields).

\textbf{References.}
Governing equations, simulation details, and parameter sweep: Appendix~C.5 of~\cite{ohana2024thewell}.

\subsection{PDE Residual Visualization}
\label{app:pde_residual}

\begin{figure}[ht]
\centering
\includegraphics[width=\textwidth]{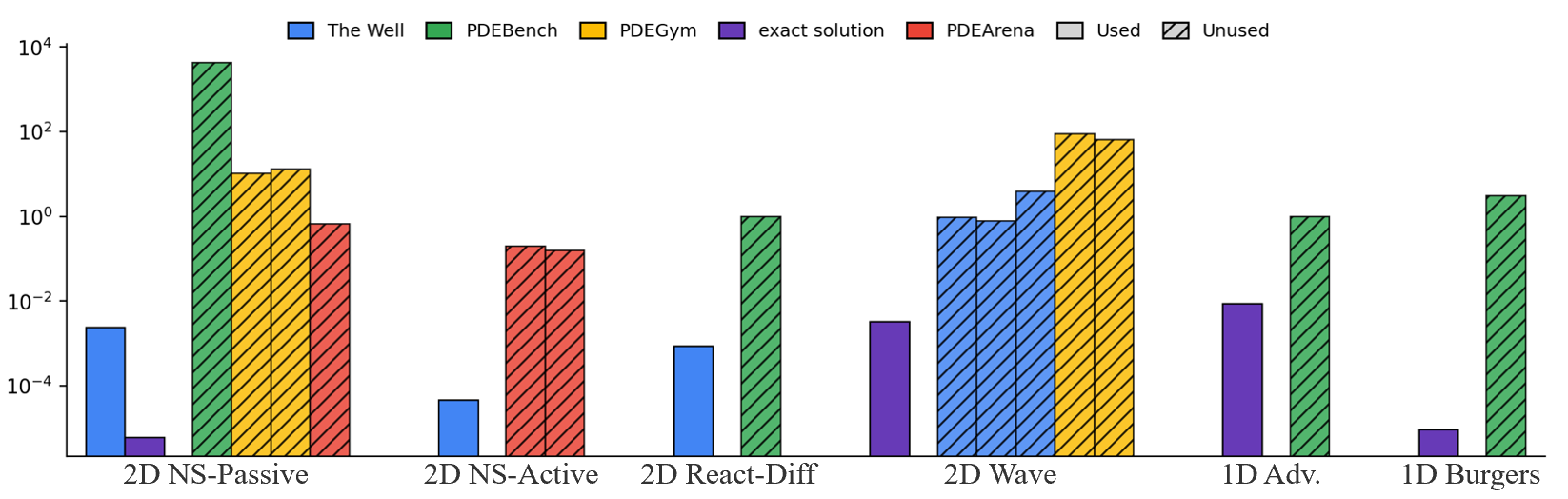}
\caption{Visualization of the discrete PDE residual computed for each downstream dataset. Exact-solution datasets yield residuals near machine precision, since their fields are evaluated analytically at every grid point. Datasets from other open-source benchmarks exhibit noticeably larger residuals than those from The Well, reflecting their numerical discretization error.}
\label{fig:pde_residual}
\end{figure}

Figure~\ref{fig:pde_residual} visualizes the discrete PDE residual computed for representative trajectories from each downstream dataset, using the same finite-difference schemes as in the UPAO. The seven exact-solution datasets yield residuals near machine precision, since their fields are evaluated analytically at every grid point. Datasets from other open-source benchmarks exhibit noticeably larger residuals than those from The Well, reflecting their numerical discretization error.

\subsection{Prediction Visualizations on The Well}
\label{app:well_predictions}

\begin{figure}[H]
\centering
\includegraphics[width=\textwidth]{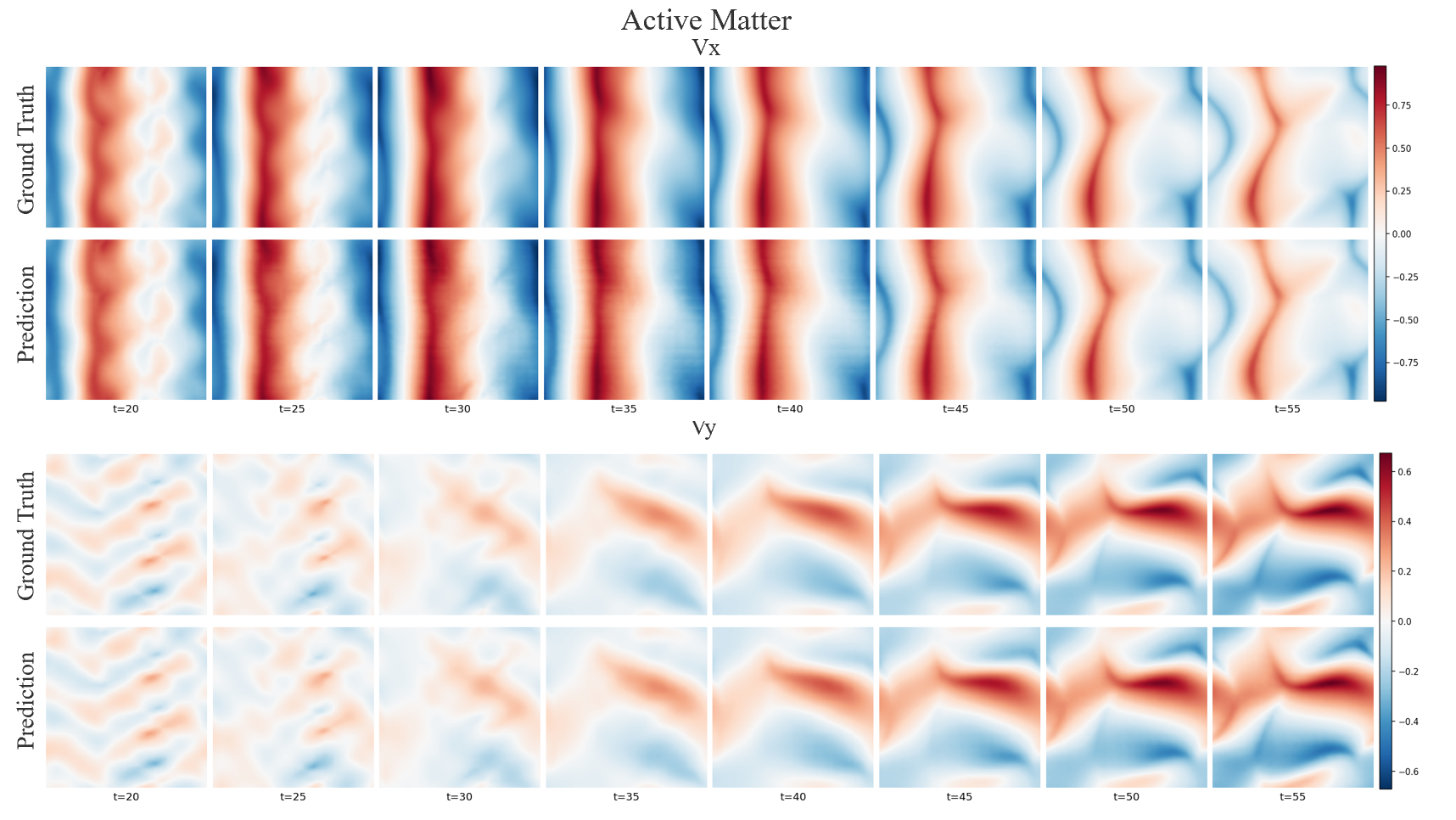}
\caption{Active Matter velocity components $V_x$ and $V_y$. Top row: ground-truth fields at two representative time steps. Bottom row: predictions from our framework after unsupervised finetuning under UPAO.}
\label{fig:pred_active_matter}
\end{figure}

\begin{figure}[H]
\centering
\includegraphics[width=\textwidth]{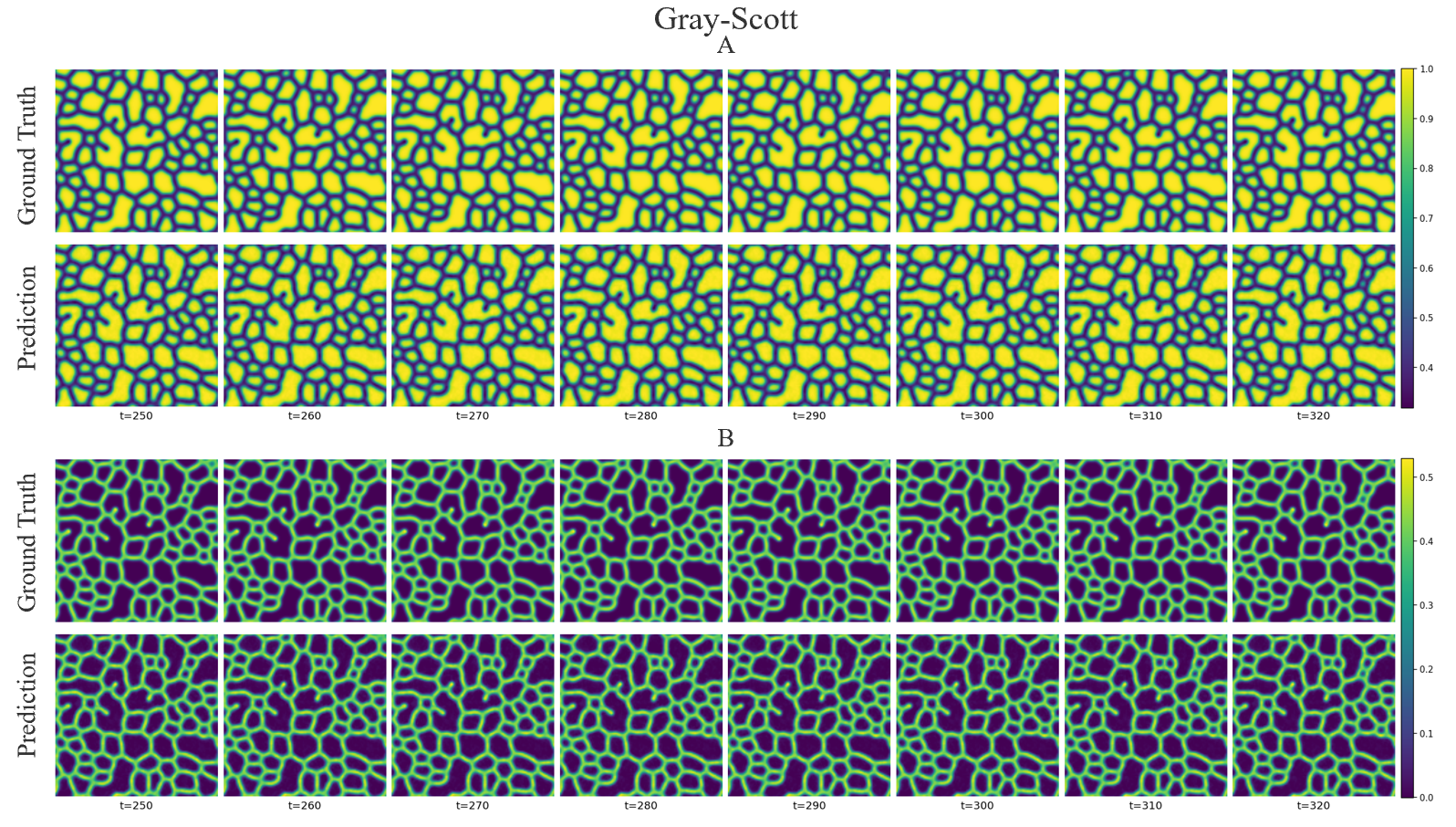}
\caption{Gray--Scott reaction-diffusion species $A$ and $B$. Top row: ground-truth fields at two representative time steps. Bottom row: predictions from our framework after unsupervised finetuning under UPAO.}
\label{fig:pred_gray_scott}
\end{figure}

\begin{figure}[H]
\centering
\includegraphics[width=\textwidth]{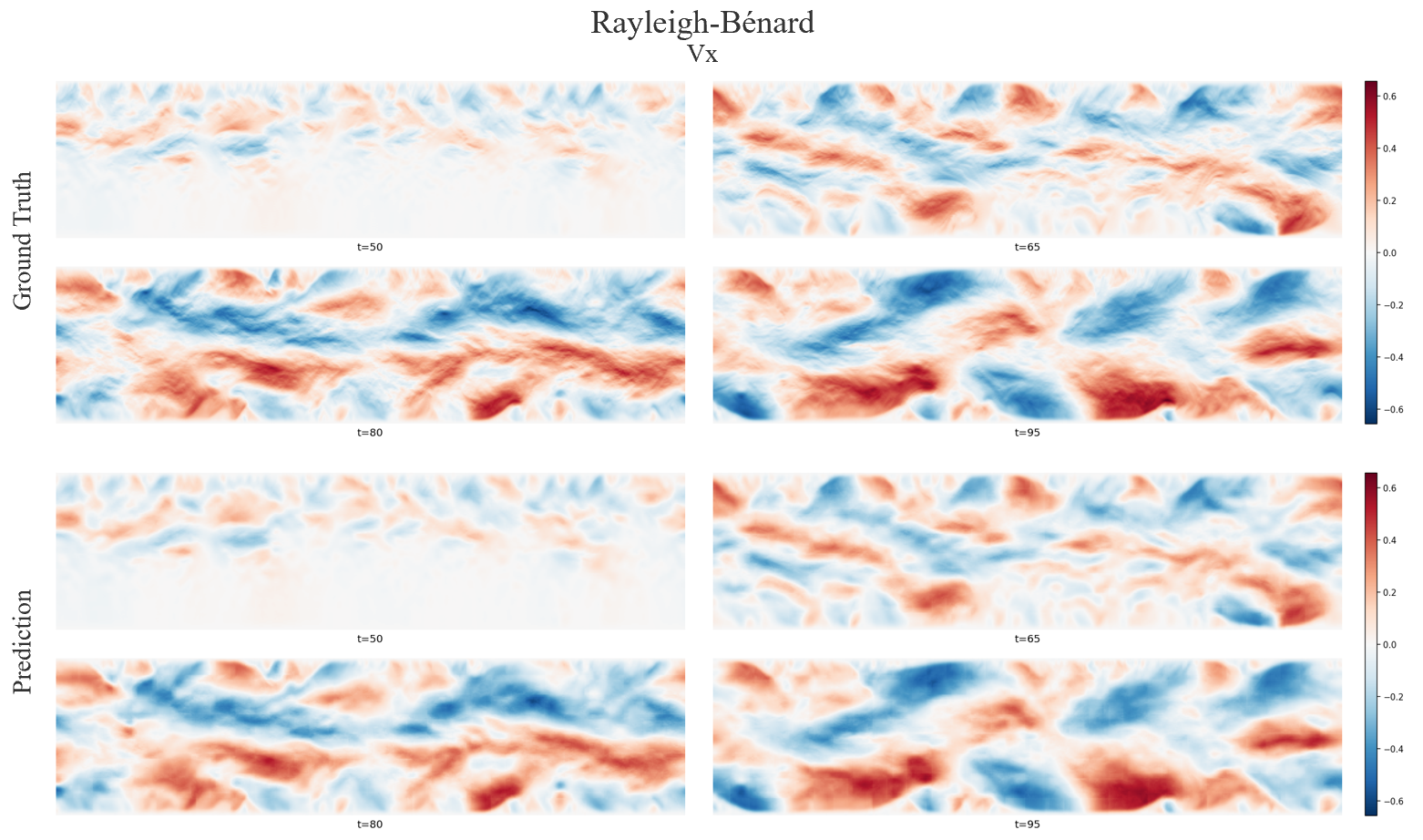}
\caption{Rayleigh--B\'enard horizontal velocity $V_x$. Top row: ground-truth field at two representative time steps. Bottom row: prediction from our framework after unsupervised finetuning under UPAO.}
\label{fig:pred_rb_vx}
\end{figure}

\begin{figure}[H]
\centering
\includegraphics[width=\textwidth]{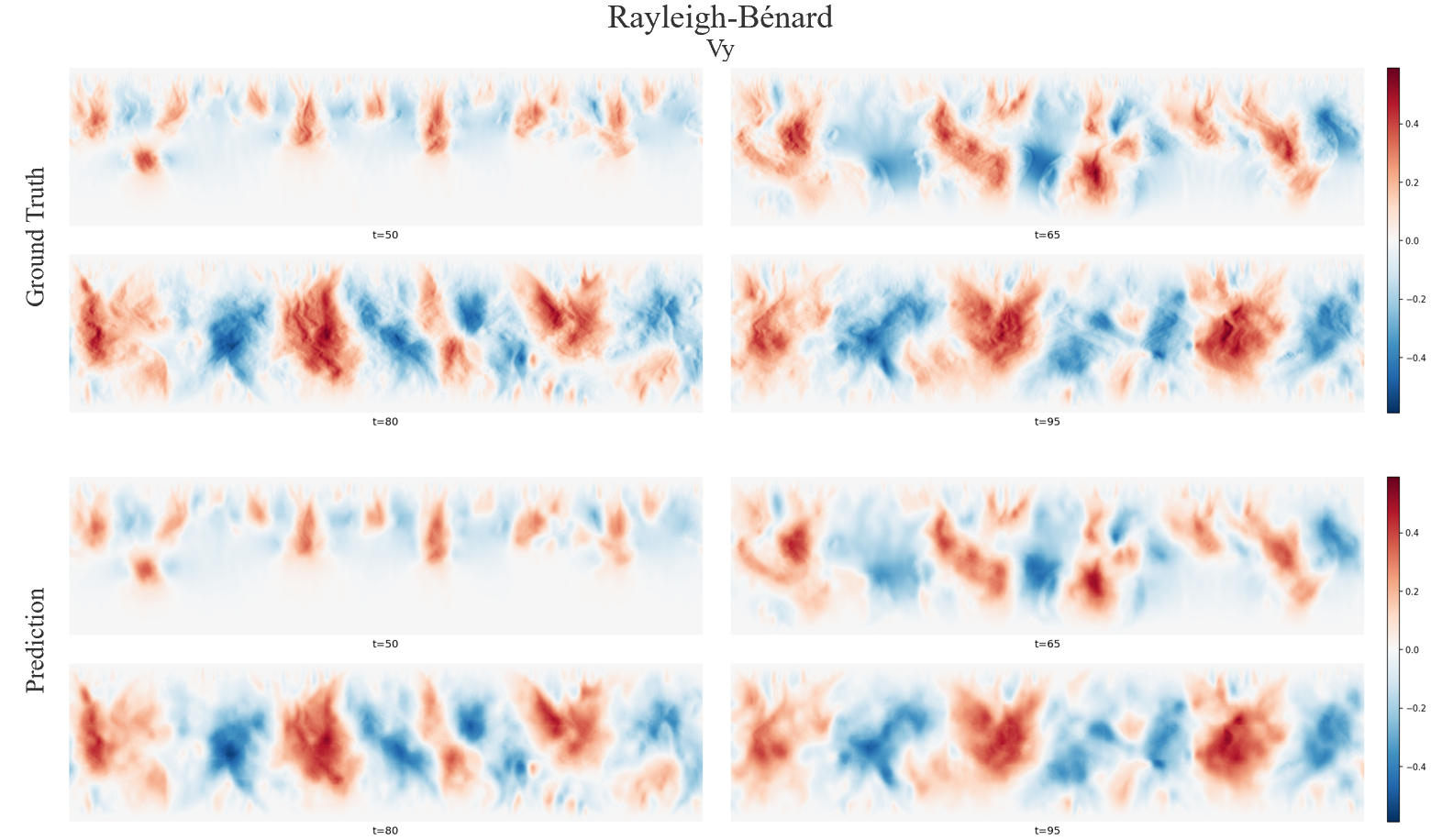}
\caption{Rayleigh--B\'enard vertical velocity $V_y$. Top row: ground-truth field at two representative time steps. Bottom row: prediction from our framework after unsupervised finetuning under UPAO.}
\label{fig:pred_rb_vy}
\end{figure}

\begin{figure}[H]
\centering
\includegraphics[width=\textwidth]{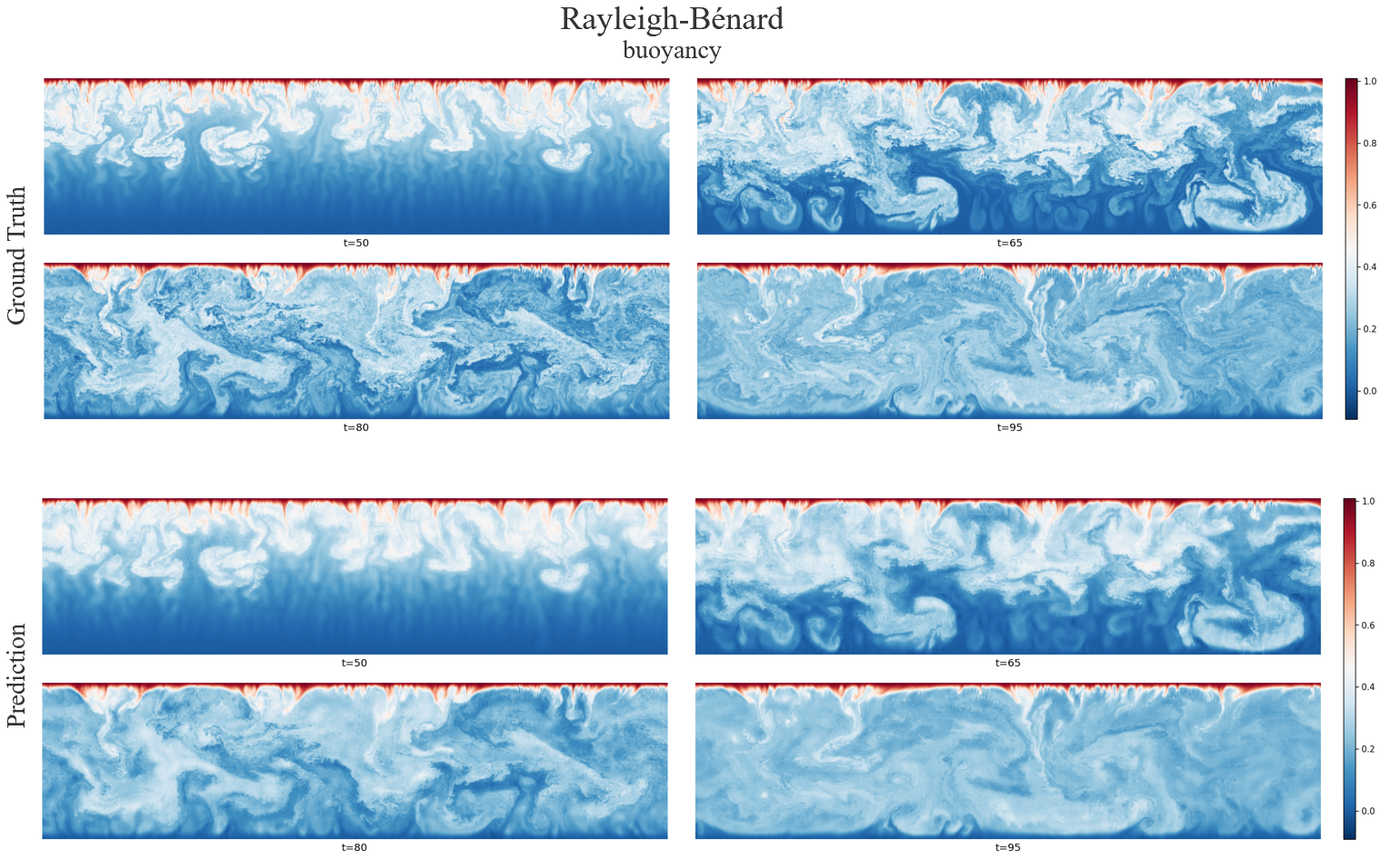}
\caption{Rayleigh--B\'enard buoyancy field. Top row: ground-truth field at two representative time steps. Bottom row: prediction from our framework after unsupervised finetuning under UPAO.}
\label{fig:pred_rb_buoy}
\end{figure}

\begin{figure}[H]
\centering
\includegraphics[width=\textwidth]{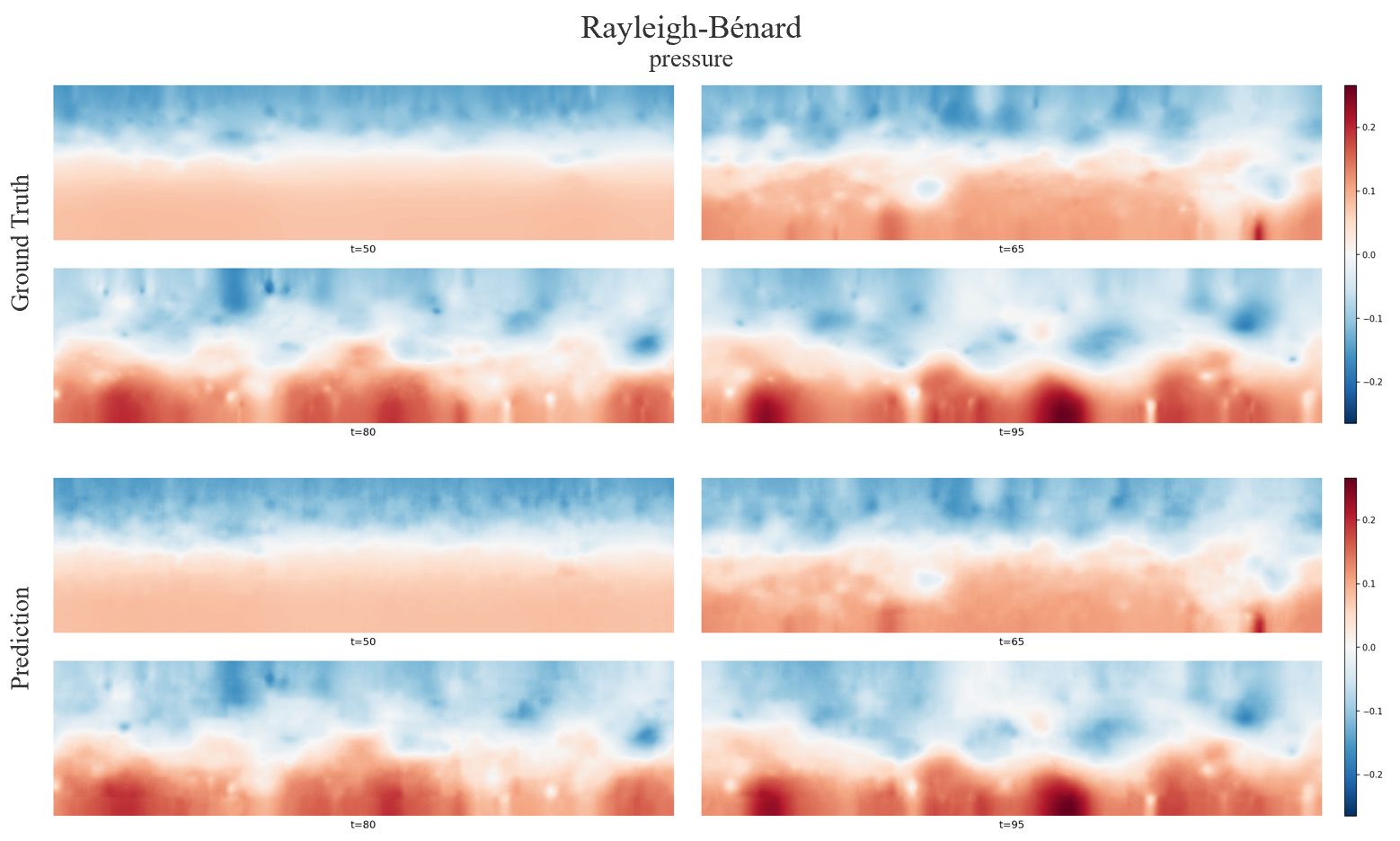}
\caption{Rayleigh--B\'enard pressure $p$. Top row: ground-truth field at two representative time steps. Bottom row: prediction from our framework after unsupervised finetuning under UPAO.}
\label{fig:pred_rb_p}
\end{figure}

\begin{figure}[H]
\centering
\includegraphics[width=\textwidth]{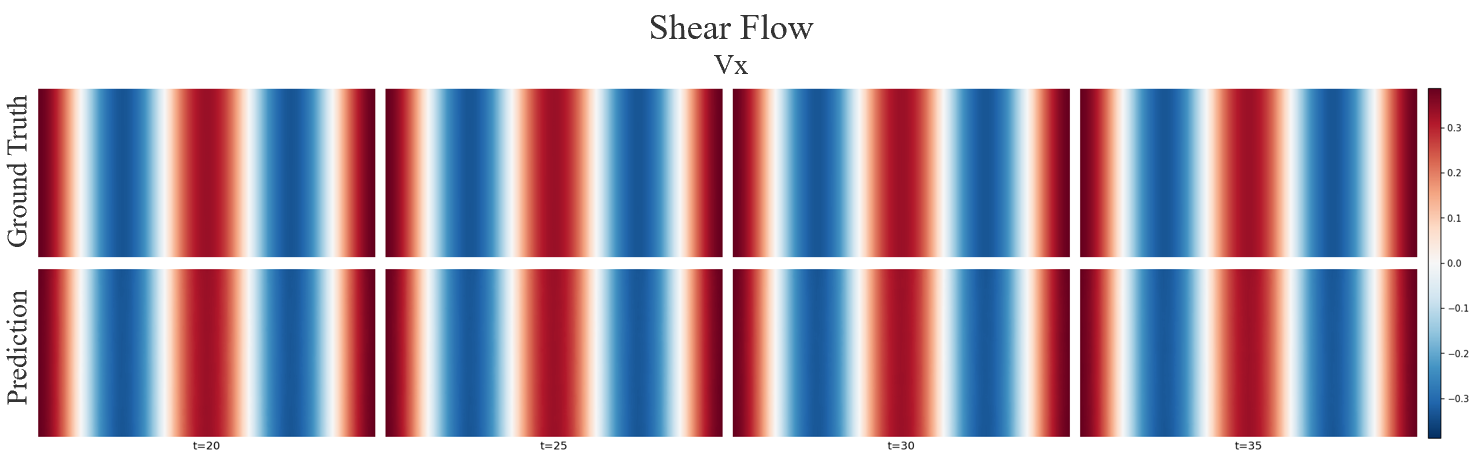}
\caption{Shear Flow horizontal velocity $V_x$. Top row: ground-truth field at two representative time steps. Bottom row: prediction from our framework after unsupervised finetuning under UPAO.}
\label{fig:pred_sf_vx}
\end{figure}

\begin{figure}[H]
\centering
\includegraphics[width=\textwidth]{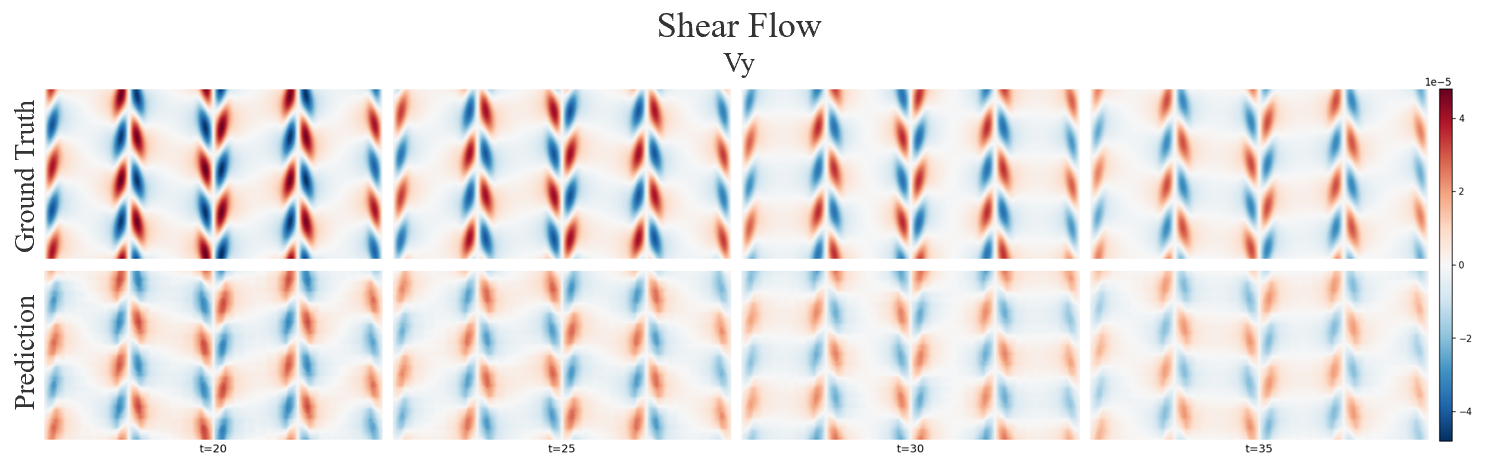}
\caption{Shear Flow vertical velocity $V_y$. Top row: ground-truth field at two representative time steps. Bottom row: prediction from our framework after unsupervised finetuning under UPAO.}
\label{fig:pred_sf_vy}
\end{figure}

\begin{figure}[H]
\centering
\includegraphics[width=\textwidth]{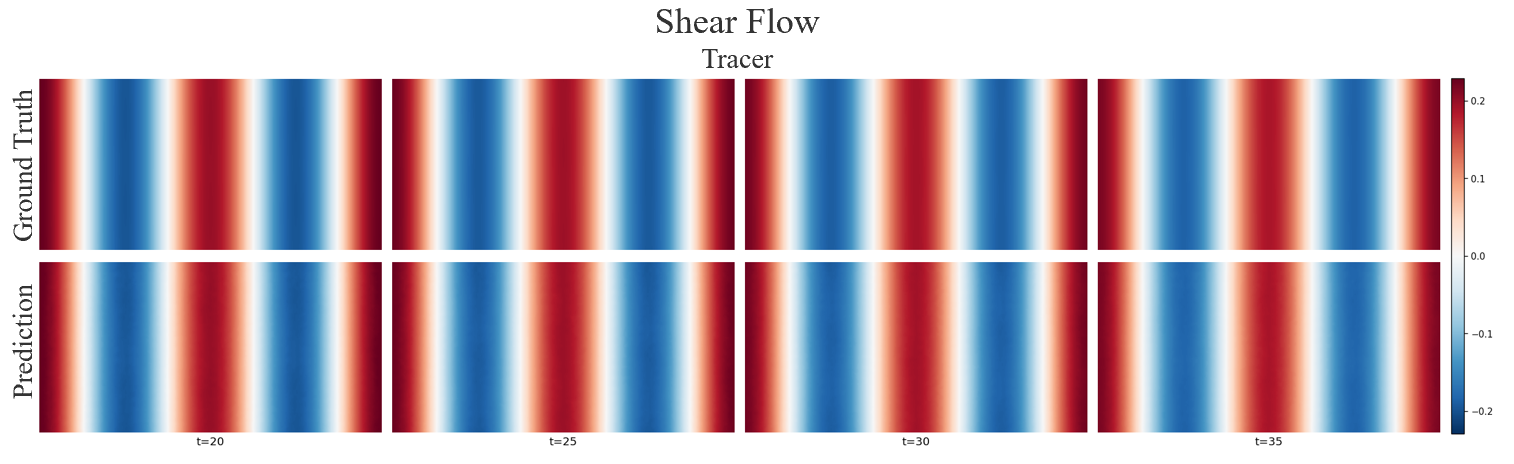}
\caption{Shear Flow passive tracer. Top row: ground-truth field at two representative time steps. Bottom row: prediction from our framework after unsupervised finetuning under UPAO.}
\label{fig:pred_sf_t}
\end{figure}

\begin{figure}[H]
\centering
\includegraphics[width=\textwidth]{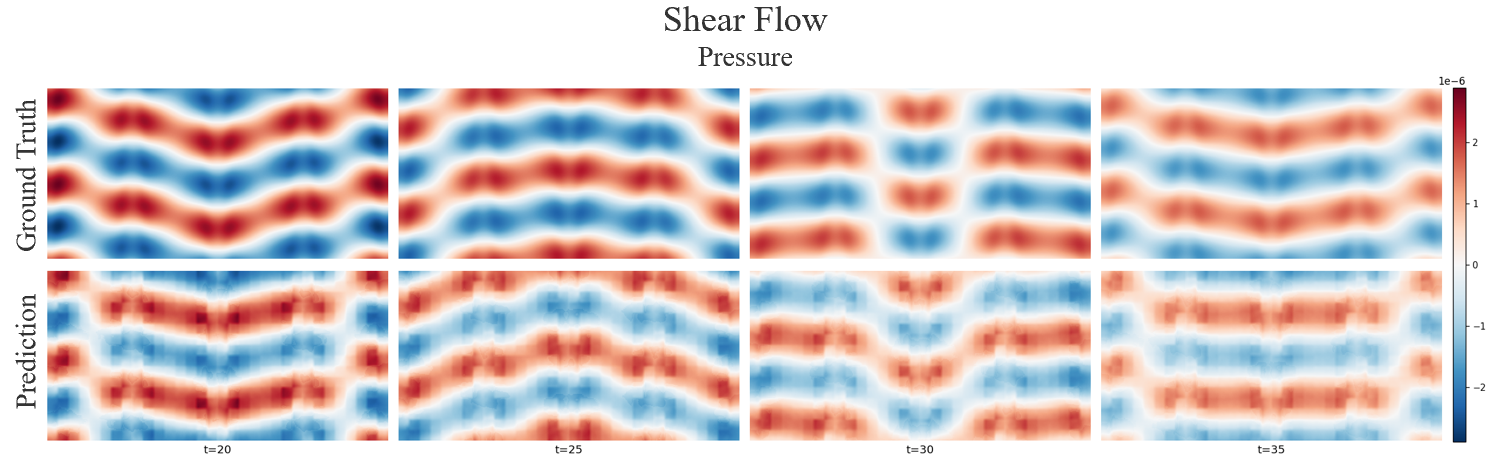}
\caption{Shear Flow pressure $p$. Top row: ground-truth field at two representative time steps. Bottom row: prediction from our framework after unsupervised finetuning under UPAO.}
\label{fig:pred_sf_p}
\end{figure}

Figures~\ref{fig:pred_active_matter} through~\ref{fig:pred_sf_p} visualize per-channel predictions on the four datasets in The Well. Each figure displays the ground-truth field on the top row and the prediction from our framework after unsupervised finetuning under UPAO on the bottom row, at two representative time steps.

\section{Baseline Configurations}
\label{app:baselines}

\subsection{Foundation-Model Baselines}
\label{app:fm_baselines}

We compare against two pretrained PDE foundation models of comparable scale to our backbone, each applied as a drop-in alternative within an identical finetuning and evaluation pipeline.

\noindent\textbf{Temporal context.} The three models do not ingest temporal context in the same way, and we state the difference explicitly because it bounds how directly the corresponding rows of Table~\ref{tab:main} can be compared. Our backbone consumes a history window of $T_{\mathrm{in}} = 8$ frames. PDE-Transformer is applied with the history length of its released ingestion protocol. Poseidon is pretrained as a continuous-in-time operator $\mathcal{G}(\mathbf{u}_0, t)$ and does not consume a history window at all. These differences are inherent to the released checkpoints rather than a choice on our part.

\textbf{PDE-Transformer (sc-b).}
We adopt the publicly released \texttt{sc-b} checkpoint at \url{https://huggingface.co/thuerey-group/pde-transformer/tree/main/sc-b}~\cite{holzschuh2025pdetransformer} and fully finetune the model for $30$ epochs on each downstream dataset. Inputs and outputs follow the original PDE-Transformer ingestion protocol, and the patch size is aligned with our backbone. We evaluate two finetuning variants. The supervised variant employs ground-truth interior fields. The unsupervised variant adopts UPAO, with no interior labels. Both variants are evaluated on the four datasets in The Well and on the four 2D exact-solution datasets at the native resolution of each dataset. The 1D and 3D exact-solution datasets fall outside the 2D training domain of the released checkpoint.

\textbf{Poseidon (B).}
We adopt the Poseidon checkpoint at \url{https://huggingface.co/camlab-ethz/Poseidon-B}~\cite{herde2024poseidon} and fully finetune for $30$ epochs. Poseidon-B accepts only square inputs. Consequently, any non-square dataset is tiled into non-overlapping $128 \times 128$ patches before finetuning and evaluation. Rayleigh--B\'enard ($512 \times 128$) yields four tiles per frame, and Shear Flow ($512 \times 256$) yields eight tiles per frame, with each tile treated as an independent sample. Beyond the spatial constraint, Poseidon-B internally FFT-resamples any input to its fixed $128\times128$ training grid, leading to severe precision degradation when evaluating finite-difference PDE residuals. In contrast, PDE-Transformer and our backbone accommodate the original resolution. We therefore restrict Poseidon-B to supervised finetuning, evaluating it on the four datasets in The Well and on the four 2D exact-solution datasets (Taylor--Green, Wave 2D, Advection--Diffusion 2D, Burgers 2D). The 1D and 3D exact-solution datasets fall outside the 2D-only input domain.

\subsection{Neural Operator Baselines}
\label{app:no_baselines}
We reproduce four competitive neural operator baselines from scratch including FNO~\cite{li2021fourier}, TFNO~\cite{kossaifi2023multigrid}, U-Net~\cite{ronneberger2015unet}, and CNextU-Net~\cite{liu2022convnext,ohana2024thewell}. For the four downstream datasets from The Well, we adopt the configurations released with the benchmark verbatim, that is, the \texttt{model} configuration files of the official codebase rather than the default settings reported in the original architecture papers. We then adapt these baselines for each exact-solution dataset by adjusting channel counts and spatial resolutions to match the target domains. All baseline models are trained for $30$ epochs from random initialization utilizing the exact same optimization and data processing pipelines as our finetuning stage. Table~\ref{tab:app_baseline_hparams} lists these hyperparameters together with the meaning each field carries in the released implementation.

\begin{table}[ht]
\centering
\caption{Hyperparameters of the four competitive neural operator baselines used across the eleven downstream datasets. All values are taken verbatim from the released configurations of the Well~\cite{ohana2024thewell}, namely \texttt{the\_well/benchmark/configs/model/\{fno,tfno,unet\_classic,unet\_convnext\}.yaml}, rather than from the default settings of the original architecture papers. ``Width'' denotes \texttt{hidden\_channels} for FNO and TFNO and \texttt{init\_features} for the two U-Nets. ``Depth'' denotes the number of Fourier blocks for FNO and TFNO, which is fixed inside the released model class and not exposed as a configuration field, and the number of encoder/decoder \texttt{stages} for the two U-Nets. For CNextU-Net, $4 \times 2$ denotes \texttt{stages}${}=4$ with \texttt{blocks\_per\_stage}${}=2$. All baselines share the AdamW optimizer and a per-GPU batch size in the listed range. Parameter counts are reported for a representative 2D downstream dataset. They vary by less than $0.1\%$ across the datasets we evaluate, since the number of active channels only affects the input and output projections.}
\label{tab:app_baseline_hparams}
\begin{tabular}{lcccccc}
\toprule
Model        & Width & Depth      & Modes          & \# Params & Optimizer (lr, wd)                                & Batch \\
\midrule
FNO          & 128   & 4          & $16 \times 16$ & 9.64M     & AdamW ($1 \times 10^{-3}$, $1 \times 10^{-4}$)    & 1--16 \\
TFNO         & 128   & 4          & $16 \times 16$ & 1.18M     & AdamW ($1 \times 10^{-3}$, $1 \times 10^{-4}$)    & 1--16 \\
U-Net        & 48    & 4          & --             & 17.46M    & AdamW ($1 \times 10^{-2}$, $1 \times 10^{-4}$)    & 1--16 \\
CNextU-Net   & 42    & 4 $\times$ 2 & --             & 18.57M    & AdamW ($1 \times 10^{-2}$, $1 \times 10^{-4}$)    & 1--16 \\
\bottomrule
\end{tabular}
\end{table}

\section{Training Protocol}
\label{app:training}

The pretraining stage employs supervised next-step prediction with a mean-squared error objective across active channels and excludes any physics-informed term, yielding a transferable representation as the backbone.

\textbf{Pretraining.}
We pretrain on the six PDEBench subsets in Section~\ref{app:pretrain} using AdamW with $\beta = (0.9, 0.95)$, learning rate $10^{-4}$ decaying to $10^{-6}$ under a cosine schedule with $200$ warmup steps, weight decay $0.01$, and gradient clipping at norm $1.0$. Training runs in full fp32 precision for $50$ epochs, with a per-GPU batch size of $24$ across $8\times$ NVIDIA A40 (46GB) under fully sharded data parallelism. Each PDEBench subset contributes $2{,}000$ clips per epoch.

\textbf{Finetuning.}
Each downstream dataset is finetuned from the pretrained backbone using AdamW with $\beta = (0.9, 0.999)$, learning rate $10^{-4}$ decaying to $10^{-6}$, $120$ warmup steps, weight decay $0.01$, and gradient clipping at norm $1.0$. Training runs in full fp32 precision for $30$ epochs, with a per-GPU batch size of $4$ on $8\times$ A40 under distributed data parallelism. The boundary weight $\lambda_{\mathrm{BC}}$ defaults to $10^4$, with dataset-specific values listed alongside the released code. The ablation in Appendix~\ref{app:lambda_bc} validates this default on the four 2D exact-solution datasets.

\section{Evaluation Protocol}
\label{app:evaluation}

\textbf{VRMSE.}
We adopt the variance-normalized root mean squared error defined in Eq.~\ref{eq:vrmse_main}. A small constant $\epsilon = 10^{-8}$ is added to the denominator for numerical stability, and inactive channels are excluded through the binary channel mask described in Appendix~\ref{app:slot_mapping}.

\textbf{Splits.}
We split each downstream dataset into $90\%$ training and $10\%$ validation by random partition of trajectories with fixed seed $42$. The validation partition serves as the evaluation set, since we do not maintain a separate held-out test partition.

\textbf{Inference.}
We do not roll out predictions autoregressively. Each prediction consumes a fixed history window of $T_{\mathrm{in}}$ frames and produces the next-frame field through a single forward pass. The reported VRMSE is the per-frame error averaged across the validation trajectories.

\section{Additional Ablations and Analyses}
\label{app:ablations}

This section provides the structural ablations and analytical experiments referenced in Section~\ref{sec:experiments}. All evaluations presented here utilize the identical pretrained backbone and apply the same UPAO framework during adaptation to ensure strict comparability.

\subsection{Boundary Weight Sensitivity}
\label{app:lambda_bc}

We adjust $\lambda_{\mathrm{BC}}$ across four orders of magnitude in Table~\ref{tab:app_lambda_bc}. $\lambda_{\mathrm{BC}}=10^4$ is a strong default. Wave 2D benefits from a stronger anchor ($10^5$), and overly large values slightly hurt TG and Burgers.

\begin{table}[ht]
\centering
\caption{Ablation study on the boundary weight $\lambda_{\mathrm{BC}}$, measured by VRMSE ($\downarrow$). Bold marks the lowest value per dataset. The default $\lambda_{\mathrm{BC}}=10^4$ is the per-column best on TG and Burgers and remains within $14\%$ of the best on AdvDiff and Wave, motivating its adoption as the default. ``---'' marks a configuration not run.}
\label{tab:app_lambda_bc}
\begin{tabular}{lcccc}
\toprule
$\lambda_{\mathrm{BC}}$ & TG 2D & Wave 2D & AdvDiff 2D & Burgers 2D \\
\midrule
$10^2$ & 0.0232 & 0.0676 & 0.0123 & 0.0092 \\
$10^3$ & 0.0170 & 0.0489 & \textbf{0.0046} & 0.0041 \\
$10^4$ (default) & \textbf{0.0070} & 0.0275 & 0.0049 & \textbf{0.0034} \\
$10^5$ & 0.0098 & \textbf{0.0243} & --- & 0.0043 \\
\bottomrule
\end{tabular}
\end{table}

\subsection{LoRA versus Full finetuning}
\label{app:lora_vs_full}

Table~\ref{tab:app_lora_vs_full} compares LoRA (rank $r\!=\!16$), which updates approximately $6.7\%$ of the total parameters, against full finetuning of the same pretrained backbone under UPAO. The two perform comparably across datasets, and LoRA matches or exceeds full finetuning on half of them, supporting the parameter-efficient design choice.

\begin{table}[ht]
\centering
\caption{Ablation study comparing LoRA against full-parameter finetuning of the same pretrained backbone, measured by VRMSE ($\downarrow$). Both variants are trained under UPAO.}
\label{tab:app_lora_vs_full}
\begin{tabular}{lcccc}
\toprule
& TG 2D & Wave 2D & AdvDiff 2D & Burgers 2D \\
\midrule
NSLoRA & 0.0060 & \textbf{0.0218} & \textbf{0.0038} & \textbf{0.0021} \\
Full finetuning & \textbf{0.0057} & 0.0252 & 0.0053 & 0.0024 \\
\bottomrule
\end{tabular}
\end{table}

\subsection{Per-Channel Results}
\label{app:per_channel}

Table~\ref{tab:app_per_channel} reports per-channel VRMSE for Standard LoRA~\cite{hu2022lora} versus NSLoRA. Both methods share the same low-rank matrices $A,B$ from the standard-LoRA warm-up and differ only in the forward path. Detailed implementations can be found in Appendix~\ref{app:nslora_impl}. The orthogonalized update helps most on the weakest channels, improving pressure on TG and both velocity components on Burgers.

\begin{table}[ht]
\centering
\caption{Complete per-channel breakdown for the controlled comparison of Table~\ref{tab:ablation_orthlora} across all eight 2D benchmarks, measured by VRMSE ($\downarrow$). Standard LoRA and NSLoRA share the same warmed-up initialization matrices $A,B$, differing exclusively in the forward path. Bold marks the lower value per channel. $\Delta$ reports the relative change of NSLoRA against Standard LoRA, computed as $(\mathrm{NSLoRA} - \mathrm{Standard\,LoRA}) / \mathrm{Standard\,LoRA} \times 100\%$, with negative values indicating that NSLoRA improves upon Standard LoRA.}
\label{tab:app_per_channel}
\begin{tabular}{l l r r r}
\toprule
Dataset & Channel & Standard LoRA & NSLoRA & $\Delta$ (\%) \\
\midrule
\multirow{3}{*}{TG 2D}              & $V_x$    & \textbf{0.00098} & 0.00102          & $+4.1$ \\
                                    & $V_y$    & \textbf{0.00098} & 0.00103          & $+5.1$ \\
                                    & $p$      & 0.01675          & \textbf{0.01582} & $-5.6$ \\
\midrule
\multirow{2}{*}{Wave 2D}            & $u$      & 0.01200          & \textbf{0.01132} & $-5.7$ \\
                                    & $w$      & 0.03284          & \textbf{0.03232} & $-1.6$ \\
\midrule
AdvDiff 2D                          & $u$      & 0.00422          & \textbf{0.00377} & $-10.7$ \\
\midrule
\multirow{2}{*}{Burgers 2D}         & $V_x$    & 0.00259          & \textbf{0.00232} & $-10.4$ \\
                                    & $V_y$    & 0.00200          & \textbf{0.00193} & $-3.5$ \\
\midrule
\multirow{2}{*}{Gray--Scott}        & $A$      & 0.09717          & \textbf{0.09697} & $-0.2$ \\
                                    & $B$      & 0.12311          & \textbf{0.12274} & $-0.3$ \\
\midrule
\multirow{2}{*}{Active Matter}      & $V_x$    & 0.12215          & \textbf{0.12025} & $-1.5$ \\
                                    & $V_y$    & 0.12769          & \textbf{0.12266} & $-3.9$ \\
\midrule
\multirow{4}{*}{Rayleigh--B\'enard} & $V_x$    & 0.19638          & \textbf{0.18939} & $-3.6$ \\
                                    & $V_y$    & 0.22663          & \textbf{0.22041} & $-2.7$ \\
                                    & buoy     & 0.22026          & \textbf{0.21610} & $-1.9$ \\
                                    & $p$      & 0.10485          & \textbf{0.09881} & $-5.8$ \\
\midrule
\multirow{4}{*}{Shear Flow}         & $V_x$    & \textbf{0.01424} & 0.01506          & $+5.8$ \\
                                    & $V_y$    & 0.25114          & \textbf{0.24663} & $-1.8$ \\
                                    & tracer   & 0.01852          & \textbf{0.01591} & $-14.1$ \\
                                    & $p$      & 0.17240          & \textbf{0.16700} & $-3.1$ \\
\bottomrule
\end{tabular}
\end{table}

\subsection{Full Cross-Target Transfer on Competitive Baselines}
\label{app:cross_domain_full}

Table~\ref{tab:app_cross_domain_full} reports the complete cross-target panels for CNextU-Net and TFNO, the two strongest competitive neural operator baselines on average across the eleven downstream datasets of Table~\ref{tab:main}. The main text Table~\ref{tab:cross_domain} shows only the CNextU-Net panels and the corresponding zero-shot and UPAO comparison on our backbone. The TFNO panels exhibit the same pattern, with both zero-shot transfer and UPAO finetuning remaining at least a factor of $10$ worse than the in-domain reference on the majority of off-diagonal cells.

\textbf{Construction of the cross-target panels.} Tables~\ref{tab:cross_domain} and~\ref{tab:app_cross_domain_full} are constructed as $7 \times 7$ panels over the $2$D downstream datasets. Each row indicates the source dataset that provides the trained checkpoint, and each column indicates the target dataset on which the model is evaluated. The diagonal reports the in-domain reference, namely the supervised checkpoint trained and evaluated on the same dataset.

\textbf{Zero-shot panel.} Off-diagonal cells in the zero-shot panel evaluate the source checkpoint on the target validation split without any further training. To address heterogeneous channel counts, we adapt the input and output projections of the source checkpoint by reshaping along the channel axis. When the source carries at least as many active channels as the target ($C_{\mathrm{src}} \geq C_{\mathrm{tgt}}$), we enumerate all $\binom{C_{\mathrm{src}}}{C_{\mathrm{tgt}}}$ channel subsets, evaluate each on the target validation split, and report the minimum VRMSE. When $C_{\mathrm{src}} < C_{\mathrm{tgt}}$, no channel subset of the source can cover the target, so the cell is marked ``---''.

\textbf{UPAO panel.} Off-diagonal cells in the UPAO panel apply the same channel adaptation and then finetune the source checkpoint on the target dataset for $5$ epochs under the unsupervised PDE-based objective with $\lambda_{\mathrm{BC}} = 10^4$. When $C_{\mathrm{src}} < C_{\mathrm{tgt}}$, the missing source channels are zero-padded so that finetuning proceeds on a uniform channel layout. The remaining hyperparameters follow Appendix~\ref{app:training}.

\begin{table}[ht]
\centering
\caption{Full cross-target VRMSE ($\downarrow$) for CNextU-Net~\cite{liu2022convnext} and TFNO~\cite{kossaifi2023multigrid} under UPAO finetuning and zero-shot inference. Each cell reports validation VRMSE on the target dataset (column) when starting from a checkpoint trained on the source dataset (row). Diagonal entries (italics) report source-only training as the in-domain reference. Bold marks the best off-diagonal value per row within each panel. ``---'' marks incompatible channel counts in the zero-shot setting. Each off-diagonal zero-shot cell reports the best VRMSE over channel-subset selections of the source. Values are reported to four decimal places. Setup matches Table~\ref{tab:cross_domain} in the main text. Dataset abbreviations: BG (Burgers 2D), WV (Wave 2D), TG (Taylor--Green 2D), AM (Active Matter), GS (Gray--Scott), RB (Rayleigh--B\'enard), SF (Shear Flow).}
\label{tab:app_cross_domain_full}
\resizebox{\textwidth}{!}{
\setlength{\tabcolsep}{3pt}
\begin{tabular}{l | ccccccc | ccccccc}
\toprule
\multicolumn{1}{l}{} & \multicolumn{14}{c}{\textbf{CNextU-Net~\cite{liu2022convnext}}} \\
\multicolumn{1}{l}{} & \multicolumn{7}{c}{\textbf{Zero-shot}} & \multicolumn{7}{c}{\textbf{UPAO}} \\
\cmidrule(lr){2-8} \cmidrule(lr){9-15}
\multicolumn{1}{l}{} & BG & WV & TG & AM & GS & RB & \multicolumn{1}{c}{SF} & BG & WV & TG & AM & GS & RB & SF \\
\midrule
BG & \textit{0.0165} & \textbf{0.2440} & --- & --- & 0.2576 & --- & --- & \textit{0.0165} & 0.3200 & 0.4098 & 1.5086 & \textbf{0.2424} & 0.5900 & 0.5383 \\
WV & \textbf{0.0845} & \textit{0.0082} & --- & --- & 0.2900 & --- & --- & \textbf{0.0200} & \textit{0.0082} & 0.3614 & 0.9588 & 0.3419 & 0.5854 & 0.6025 \\
TG & 0.3129 & 0.3522 & \textit{0.0178} & 0.8263 & \textbf{0.2918} & --- & --- & 0.1960 & 0.3582 & \textit{0.0178} & 0.6888 & \textbf{0.1456} & 0.5779 & 0.4282 \\
AM & 0.0555 & 0.0676 & \textbf{0.0485} & \textit{0.1248} & 0.2894 & --- & --- & \textbf{0.0367} & 0.1769 & 0.0474 & \textit{0.1248} & 0.2188 & 0.4404 & 0.4802 \\
GS & 0.4596 & \textbf{0.3965} & --- & --- & \textit{0.0054} & --- & --- & 0.2828 & \textbf{0.1657} & 0.3729 & 0.9479 & \textit{0.0054} & 0.6455 & 0.5094 \\
RB & \textbf{0.1197} & 0.1550 & 0.1561 & 0.3185 & 0.4879 & \textit{0.1022} & 0.2017 & \textbf{0.0604} & 0.1314 & 0.0624 & 0.3673 & 0.2701 & \textit{0.1022} & 0.2318 \\
SF & \textbf{0.0629} & 0.2403 & 0.0635 & 0.7857 & 0.2486 & 0.4492 & \textit{0.0455} & 0.2094 & 0.3964 & \textbf{0.1703} & 0.8347 & 0.5213 & 0.6027 & \textit{0.0455} \\
\bottomrule
\end{tabular}
}

\vspace{0.8em}

\resizebox{\textwidth}{!}{
\setlength{\tabcolsep}{3pt}
\begin{tabular}{l | ccccccc | ccccccc}
\toprule
\multicolumn{1}{l}{} & \multicolumn{14}{c}{\textbf{TFNO~\cite{kossaifi2023multigrid}}} \\
\multicolumn{1}{l}{} & \multicolumn{7}{c}{\textbf{Zero-shot}} & \multicolumn{7}{c}{\textbf{UPAO}} \\
\cmidrule(lr){2-8} \cmidrule(lr){9-15}
\multicolumn{1}{l}{} & BG & WV & TG & AM & GS & RB & \multicolumn{1}{c}{SF} & BG & WV & TG & AM & GS & RB & SF \\
\midrule
BG & \textit{0.1615} & \textbf{0.9080} & --- & --- & 0.9947 & --- & --- & \textit{0.1615} & 0.3593 & 0.2463 & 1.6661 & \textbf{0.0243} & 0.4449 & 0.5201 \\
WV & \textbf{0.1246} & \textit{0.0074} & --- & --- & 0.2721 & --- & --- & 0.0241 & \textit{0.0074} & 0.2553 & 0.8399 & \textbf{0.0062} & 0.3253 & 0.4204 \\
TG & \textbf{0.9364} & 0.9511 & \textit{0.0033} & 0.9632 & 0.9840 & --- & --- & 0.1745 & 0.3956 & \textit{0.0033} & 0.5088 & \textbf{0.0227} & 0.5002 & 0.5663 \\
AM & 0.1187 & 0.2663 & \textbf{0.0861} & \textit{0.1390} & 0.5361 & --- & --- & 0.0410 & 0.0939 & 0.0192 & \textit{0.1390} & \textbf{0.0059} & 0.3289 & 0.3987 \\
GS & 0.3999 & \textbf{0.3663} & --- & --- & \textit{0.0033} & --- & --- & 0.0764 & \textbf{0.0448} & 0.3790 & 0.7755 & \textit{0.0033} & 0.3204 & 0.4958 \\
RB & 0.3327 & 0.2266 & \textbf{0.2148} & 0.4710 & 0.4083 & \textit{0.1762} & 0.4499 & 0.0829 & 0.0958 & 0.0781 & 0.2475 & \textbf{0.0068} & \textit{0.1762} & 0.2752 \\
SF & \textbf{0.5410} & 0.5808 & 0.7065 & 0.7893 & 0.5601 & 0.7742 & \textit{0.2234} & 0.1149 & 0.1709 & 0.2132 & 0.4788 & \textbf{0.0082} & 0.3268 & \textit{0.2234} \\
\bottomrule
\end{tabular}
}
\end{table}

\subsection{Newton-Schulz vs SVD Speed Comparison}
\label{app:ns_vs_svd}

We compare the speed of two orthogonalization paths for the LoRA matrices $A$ and $B$: the 5-step Newton-Schulz iteration of NSLoRA, and an explicit SVD computation that returns the exact polar factor $U V^\top$ via \texttt{torch.linalg.svd}. The two paths are not exactly equivalent. With the coefficients of~\cite{jordan2024muon}, which are tuned for throughput rather than for convergence, $\mathrm{NS}_5$ returns a matrix whose nontrivial singular values lie in a bounded band around one rather than exactly at one, as stated in Appendix~\ref{app:nslora_impl}. NSLoRA does not require exact orthogonality, since the mechanism it relies on is the suppression of near-zero singular values, which the bounded band already provides. Both paths use the same real LoRA matrices loaded from a Shear Flow init checkpoint ($108$ $(A, B)$ pairs across the five module types $\{\mathrm{qkv}, \mathrm{proj}, \mathrm{gate\_proj}, \mathrm{up\_proj}, \mathrm{down\_proj}\}$) and run on a single NVIDIA A100-SXM4-40GB under identical warm-up and synchronization. Reported values are the mean per-call time across $T = 5$ independent trials, with the standard deviation across trials reported as $\pm$ ($1\sigma$). The full LoRA-layer forward timings in Table~\ref{tab:app_ns_layer} average $K = 200$ iterations per trial.

Summing the execution times across a $12$-block transformer with $5$ LoRA modules per block demonstrates an end-to-end forward pass of $158.19$ ms for Newton-Schulz versus $217.75$ ms for SVD, yielding a $1.38\times$ overall speedup. Newton-Schulz circumvents the kernel invocation overhead of cuSOLVER eigendecomposition routines by relying on highly optimized matrix multiplications. 

\begin{table}[ht]
\centering
\caption{Per-layer forward speed within the SplitNS-style LoRA layer at batch size $8$ and sequence length $1024$, in milliseconds per call (mean $\pm 1\sigma$ across $5$ trials, $K = 200$ iterations per trial). Bold marks the largest NS speedup.}
\label{tab:app_ns_layer}
\begin{tabular}{llrrr}
\toprule
Module & in $\to$ out & NS forward (ms) & SVD forward (ms) & NS speedup \\
\midrule
qkv       & $768 \to 2304$ & $2.5023 \pm 0.0003$ & $3.2566 \pm 0.0003$ & $1.30\times$ \\
proj      & $768 \to 768$  & $1.4137 \pm 0.0001$ & $2.1249 \pm 0.0005$ & $\mathbf{1.50\times}$ \\
gate\_proj & $768 \to 3072$ & $3.1113 \pm 0.0003$ & $4.2344 \pm 0.0004$ & $1.36\times$ \\
up\_proj   & $768 \to 3072$ & $3.1084 \pm 0.0001$ & $4.2955 \pm 0.0005$ & $1.38\times$ \\
down\_proj & $3072 \to 768$ & $3.0471 \pm 0.0001$ & $4.2347 \pm 0.0002$ & $1.39\times$ \\
\bottomrule
\end{tabular}
\end{table}

\newpage

\end{document}